\documentclass[runningheads]{llncs}

\usepackage{eccv}

\usepackage{eccvabbrv}

\usepackage{graphicx}
\usepackage{booktabs}
\usepackage{wrapfig}
\usepackage[accsupp]{axessibility}  

\usepackage{hyperref}

\usepackage{orcidlink}
\usepackage{booktabs}
\usepackage[table]{xcolor} 
\usepackage{multirow}
\usepackage{bm}

\makeatletter
\newcommand{\printfnsymbol}[1]{%
  \textsuperscript{\@fnsymbol{#1}}%
}
\makeatother

\begin{document}

\title{MVFusion-GS: Motion-Variance Guided Temporal Attention for High-Quality Dynamic Gaussian Splatting} 

\titlerunning{MVFusion-GS for High-Quality Dynamic Gaussian Splatting}

\author{
Jianwei Hu\inst{1}\textsuperscript{*} \and
Tingxuan Huang\inst{1}\textsuperscript{*} \and
Hengyu Zhou\inst{1} \and
Ningna Wang\inst{2} \and
Xiaohu Guo\inst{2} \and
Jinshan Lai\inst{3} \and
Bin Wang\inst{1}\textsuperscript{\ensuremath{\dagger}}
}

\authorrunning{Jianwei Hu{*}, Tingxuan Huang{*}, et al.}

\institute{Tsinghua University, Beijing, China
\and
The University of Texas at Dallas, Richardson, TX, USA
\and
University of Electronic Science and Technology of China, Chengdu, China\\[1mm]
{\scriptsize
\email{
hjw17@tsinghua.org.cn,
\{htx25,zhouhy22\}@mails.tsinghua.edu.cn,
\{ningna.wang,xguo\}@utdallas.edu,
laijinshan574@gmail.com,
wangbins@tsinghua.edu.cn
}
}
}

\maketitle

\begingroup
\renewcommand{\thefootnote}{*}
\footnotetext{Equal contribution.}
\renewcommand{\thefootnote}{\ensuremath{\dagger}}
\footnotetext{Corresponding author.}
\endgroup


\begin{figure*}[ht]
    \centering
    \includegraphics[width=0.95\linewidth]{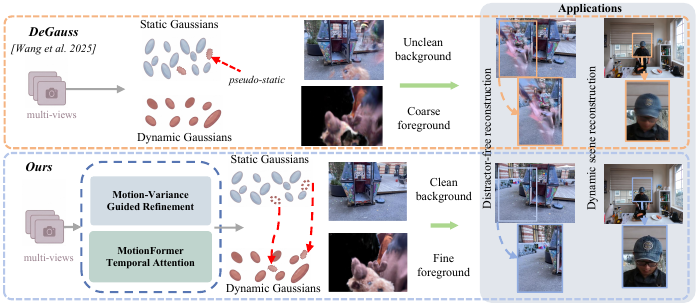}
    \caption{Standard dynamic-static Gaussian decomposition leaves pseudo-static foreground residuals in the static branch (top), causing background contamination. Our method enhances the motion awareness of the deformation network, correctly re-attributing pseudo-static Gaussians to the dynamic branch (bottom). This yields dual benefits: cleaner backgrounds for distractor-free reconstruction and finer foreground details for dynamic scene reconstruction (right).
}
    \label{fig:teaser}
\end{figure*}

\begin{abstract}
\vspace{-20pt}
3D Gaussian Splatting (3DGS) enables real-time novel view synthesis for static scenes. Extending it to dynamic scenes via deformation fields has recently attracted significant attention, particularly for \textit{dynamic scene reconstruction} and \textit{distractor-free reconstruction}. However, existing deformation networks lack explicit motion awareness: they neither capture long-term motion intensity nor exploit short-term temporal coherence, leading to inaccurate foreground deformation and pseudo-static residuals in the background. We present \textbf{MVFusion-GS}, a method that enhances deformation networks with two complementary motion-aware mechanisms. The \textbf{Motion-Variance Guided Refinement} aggregates per-Gaussian deformation statistics across time to estimate motion variance and uses it to guide dynamic-static separation during deformation prediction. The \textbf{MotionFormer Temporal Attention} module applies Transformer self-attention over neighboring timesteps to model local motion dependencies and improve temporal consistency. Extensive experiments on both dynamic scene reconstruction and distractor-free reconstruction benchmarks demonstrate state-of-the-art performance, showing that explicit motion awareness improves both foreground motion modeling and static background reconstruction.
\end{abstract}

\section{Introduction}
\label{sec:intro}

Extending 3D Gaussian Splatting (3DGS)~\cite{3dgs} to dynamic scenes via deformation fields is a central research topic, where two representative tasks are \textit{dynamic scene reconstruction} and \textit{distractor-free reconstruction}. The former aims to faithfully reconstruct foreground motion with high deformation fidelity~\cite{deformablegs,4dgs,scgs}, while the latter seeks to recover a clean static background by separating out transient distractors (e.g., pedestrians, vehicles)~\cite{wildgaussians,spotlesssplats}. Both tasks rely on deformation fields that map Gaussian attributes from a canonical space to each observation timestep, yet their optimization objectives diverge: one pursues high-fidelity foreground fitting, the other clean background separation. Simultaneously addressing both within a single framework remains challenging.

DeGauss~\cite{degauss} tackles this with a decoupled dynamic-static Gaussian framework, decomposing the scene into foreground dynamic Gaussians modeled by a HexPlane-based spatio-temporal deformation module and background static Gaussians that directly fit the scene structure, composing them via a learnable probabilistic mask. While effective for distractor-free reconstruction, its deformation network lacks motion-aware cues. Fig.~\ref{fig:teaser} shows that when foreground objects exhibit low-amplitude or transient motion, the network fails to produce accurate deformation predictions, leaving pseudo-static foreground residuals in the background branch and degrading static reconstruction quality.

The root cause is that standard deformation networks encode latent features through spatio-temporal grids without explicitly modeling the motion pattern each Gaussian undergoes or its temporal context across adjacent frames. To address this, we propose two complementary mechanisms:

\textbf{Motion-Variance Guided Refinement.}
We periodically sample each Gaussian's deformation across multiple timesteps and compute statistics over its positional, rotational, and scale variations, yielding a 13D global trajectory signature. Injected into the deformation network's latent space, this signature provides an explicit motion-intensity prior for foreground-background discrimination. 

\textbf{MotionFormer Temporal Attention.}
Complementing the global trajectory signature, we introduce a Transformer-based temporal attention module that aggregates neighboring timesteps through query-centered cross-attention, enabling the network to leverage local temporal context for temporally consistent deformation prediction.

The two mechanisms work synergistically: the global trajectory signature provides long-term motion priors for coarse dynamic-static separation, while temporal attention refines instantaneous deformation through short-term temporal context. This yields dual benefits --- more accurate foreground motion capture in dynamic scene reconstruction, and cleaner background purity in distractor-free reconstruction by correctly re-attributing pseudo-static Gaussians to the dynamic branch. 

Our main contributions are:
\begin{itemize}
\item A dual-task plug-in module that enhances deformation network motion awareness, achieving consistent improvements on both dynamic scene reconstruction and distractor-free reconstruction benchmarks.
\item Motion-Variance Guided Refinement, which aggregates per-Gaussian deformation statistics across the global time span into an explicit motion intensity prior for dynamic--static separation.
\item MotionFormer Temporal Attention, which leverages cross-attention to model cross-temporal motion dependencies, improving deformation accuracy and temporal consistency.

\noindent Code is available at: \url{https://github.com/toseeai-com/MVFusion-GS}.
\end{itemize}

\section{Related Work}
\label{sec:related_works}

\subsection{Novel View Synthesis for Static Scenes}

Neural Radiance Fields (NeRF)~\cite{nerf} introduced continuous volumetric scene representations for novel view synthesis, with later works improving efficiency~\cite{mobilenerf, instantngp} and geometric accuracy~\cite{deepsdf, volsdf}. More recently, 3D Gaussian Splatting (3DGS)~\cite{3dgs} represents scenes as anisotropic Gaussian primitives and enables real-time rendering via differentiable rasterization. Follow-up works such as Scaffold-GS~\cite{scaffoldgs} and 2DGS~\cite{2dgs} further improve scalability and rendering quality. Our method builds upon the 3DGS representation.

\subsection{Dynamic Gaussian Scene Reconstruction}

Recent works extend 3DGS to dynamic scenes by learning deformation fields that map Gaussians from a canonical space to each timestep. Deformable-GS~\cite{deformablegs} uses an MLP-based deformation network, while 4DGS~\cite{4dgs} adopts a HexPlane encoder with MLP decoders for time-dependent attribute prediction. SC-GS~\cite{scgs} introduces sparse control nodes to guide Gaussian deformation. These approaches implicitly learn motion through deformation but lack explicit awareness of the motion patterns associated with each Gaussian. 
Other recent works improve dynamic Gaussian representations from complementary perspectives. Some parameterize Gaussian motion with explicit trajectories, such as Shape-of-Motion~\cite{wang2025shapeofmotion}, SplineGS~\cite{park2025splinegs}, and Gaussian Marbles~\cite{stearns2024marbles}, which are primarily designed for monocular dynamic reconstruction. SpaceTimeGS~\cite{spacetimegs} and ST-4DGS~\cite{st4dgs} improve spatial-temporal consistency through spacetime Gaussians and motion-aware regularization, and FreeTimeGS~\cite{freetimegs} allows Gaussian primitives to appear at arbitrary times and locations for complex motion. In contrast, our method injects motion-aware signals into deformation-based Gaussian pipelines by incorporating global trajectory statistics and local temporal attention into the deformation feature space.

\subsection{Distractor Handling in Dynamic Scene Reconstruction}
Dynamic scene reconstruction from casually captured imagery is often affected by transient distractors. Some approaches suppress their influence without explicitly modeling them. RobustNeRF~\cite{sabour2023robustnerf} down-weights high-residual pixels during optimization. NeRF On-the-go~\cite{nerfonthego} predicts per-pixel dynamic masks using DINOv2 features, later extended to Gaussian Splatting in WildGaussians~\cite{wildgaussians}. SpotlessSplats~\cite{spotlesssplats} further leverages clustered diffusion features to identify transient regions. 

Another line of work explicitly models transient content. NeRF-W~\cite{martinbrualla2021nerfw} decomposes scenes into static and transient neural fields, while D\textsuperscript{2}NeRF~\cite{d2nerf} improves decomposition stability through assignment regularization. DeSplat~\cite{Wang2024DeSplatDG} extends this idea to Gaussian Splatting using view-specific Gaussian primitives to represent transient occluders. DeGauss~\cite{degauss} later proposes a dual-branch Gaussian Splatting framework with a learnable foreground–background mask. However, lacking motion-aware cues, its deformation network often misclassifies subtle or transient motion as static. Our method introduces motion trajectory priors and temporal attention to improve separation and reconstruction.
\section{Preliminary}
\subsection{3D Gaussian Splatting}

3D Gaussian Splatting (3DGS)~\cite{3dgs} represents a scene as a set of anisotropic Gaussian primitives, each defined by a mean position \(\mu\), opacity \(\alpha\), color \(c\), and covariance \(\Sigma\). Instead of storing \(\Sigma\) directly, 3DGS parameterizes it using a rotation–scale decomposition:  
\begin{equation}
\Sigma = R S S^\top R^\top,  
\end{equation}
where \(R \in \mathrm{SO}(3)\) encodes orientation, and \(S = \mathrm{diag}(s_x, s_y, s_z)\) controls the anisotropic scale along local axes. This formulation guarantees positive semi-definiteness and enables independent optimization of shape and orientation.

When rendering, each 3D Gaussian is projected to the image plane through the camera transformation \(W\) and the projection Jacobian \(J\). The resulting 2D covariance is computed as  
\begin{equation}  
\Sigma_{2D} = J W \Sigma W^\top J^\top,  
\end{equation}
which determines the ellipse footprint of the projected Gaussian. The splats are then alpha-composited in front-to-back order to produce the final image.

\subsection{Dynamic 3DGS}

Dynamic 3D Gaussian Splatting extends static 3DGS to videos by introducing a time-conditioned deformation model that maps Gaussians from a canonical space to observation space at each timestamp. Early methods (e.g., 4DGS~\cite{4dgs}, SC-GS~\cite{scgs}) typically use a single deformation network for all Gaussians, which can lead to unstable optimization: dynamic residuals may leak into static structures, while background regions may become over-deformed.

\noindent\textbf{Decoupled Dynamic--Static Representation and Composition.} To alleviate this issue, DeGauss~\cite{degauss} adopts a decoupled formulation with two Gaussian sets: a static background set $\mathcal{G}_s$ and a dynamic foreground set $\mathcal{G}_d$ whose attributes are modulated by a deformation network.

At time $t$, only the dynamic branch is deformed:
\begin{equation}
\mathcal{G}_d(t)
=
\mathcal{G}_d
+
\Delta\mathcal{G}_d(t),
\qquad
\Delta\mathcal{G}_d(t)=\Phi(h_t),
\end{equation}
where $h_t=f(\mathbf{x},t)$ denotes the spatiotemporal feature extracted from Gaussian coordinates $\mathbf{x}$ and timestamp $t$, and $\Phi(\cdot)$ predicts Gaussian attribute offsets.

Both branches are rendered independently to produce dynamic and static images $\hat{I}_d(t)$ and $\hat{I}_s(t)$. A per-pixel composition mask $M(t)\in[0,1]$ predicted from the dynamic branch blends the two renders:
\begin{equation}
\hat{I}(t)
=
M(t)\odot \hat{I}_d(t)
+
\bigl(1-M(t)\bigr)\odot \hat{I}_s(t).
\label{eq:degauss_comp}
\end{equation}

This decoupled formulation encourages motion to be explained by $\mathcal{G}_d$ while keeping the background $\mathcal{G}_s$ stable.

\section{Method}
\label{sec:method}

\subsection{Overview}
\label{sec:overview}

\begin{figure}[t]
\centering
\includegraphics[width=\linewidth]{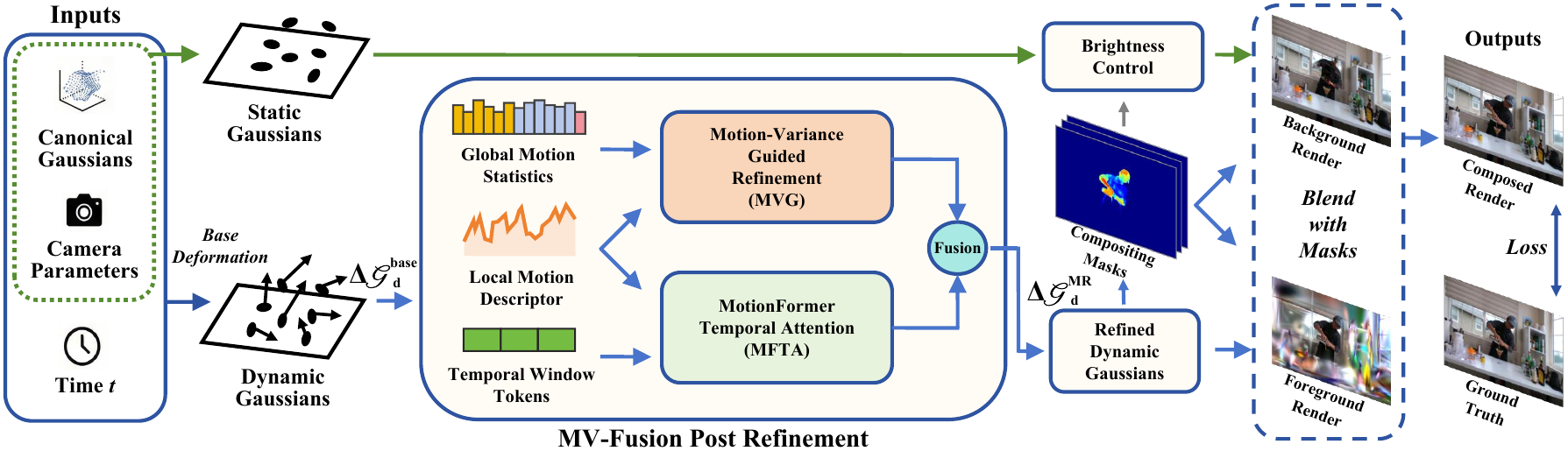}
\caption{
Overview of the proposed motion-aware refinement framework.
The baseline deformation network first predicts a coarse deformation for dynamic Gaussians.
Two motion-aware modules are then applied: MVG extracts global trajectory signatures and local variance cues, while MFTA aggregates short-term temporal context.
The fused motion-aware feature produces refined Gaussian updates, followed by foreground–background rendering and mask-based composition.
}
\label{fig:pipeline}
\end{figure}

\noindent\textbf{Motivation.}
In a decoupled dynamic--static reconstruction pipeline, the dynamic branch is responsible for modeling scene motion while keeping the static branch stable. However, standard spatiotemporal deformation networks often underfit subtle or transient motions, causing dynamic residuals to leak into the static branch and degrade the reconstruction quality.

To address this issue, we treat the baseline deformation network as a coarse predictor and introduce a lightweight motion-aware refinement. The refinement leverages motion statistics and short-term temporal context to enhance the deformation feature, enabling more accurate modeling of subtle dynamic behaviors. 

Fig.~\ref{fig:pipeline} illustrates the overall pipeline of the proposed method. The refined deformation is formulated as

\begin{equation}
\Delta \mathcal{G}_d(t)
=
\Delta \mathcal{G}_d^{\text{base}}(t)
+
\Delta \mathcal{G}_d^{\text{MR}}(t),
\end{equation}

where the motion-aware refinement term is predicted from fused features:

\begin{equation}
\Delta \mathcal{G}_d^{\text{MR}}(t)
=
\Phi\!\big(
h_{\text{base}}(t)
+
h_{\text{MVG}}(t)
+
h_{\text{MFTA}}(t)
\big),
\end{equation}

Here $h_{\text{base}}$ denotes the baseline deformation feature, and $\Phi(\cdot)$ denotes the deformation heads applied on the fused feature. In other words, MVG and MFTA operate purely in the feature space: they are fused with the baseline deformation feature and decoded by the unchanged deformation heads $\Phi$ to predict refined Gaussian attribute updates. This plug-in design requires no modification to the original deformation heads and thus integrates seamlessly into existing deformation-based Gaussian pipelines.

\noindent\textbf{Decoupled rendering.}
The refined deformation $\Delta\mathcal{G}_d(t)$ is applied to update the dynamic Gaussians. 
The decoupled Gaussian sets are rendered independently and composited using a learned soft mask $M(t)$ derived from deformation SH\textsubscript{mask} channels, which represent static, dynamic, and brightness-control components.

\subsection{Motion-Variance Guided Refinement (MVG)}
\label{sec:motion_stats}

\noindent\textbf{Motivation.}
In a decoupled dynamic--static pipeline, the deformation branch must determine which Gaussians belong to the moving foreground and how they evolve over time. However, standard spatiotemporal grids only encode local $(\mathbf{x},t)$ context and provide no explicit cue about the \emph{global motion behavior} of each Gaussian. Consequently, low-amplitude or transient motions are often under-modeled, producing ``pseudo-static'' residuals that leak into the background branch.

To address this issue, we periodically extract motion descriptors from simulated deformation trajectories and use them as motion priors. MVG contains two complementary signals:
(i) a \emph{global trajectory signature} that captures long-term motion statistics, represented as $\mathbf{V}\in\mathbb{R}^{N\times13}$; and
(ii) a cached \emph{local variance dictionary} that maps a query timestamp $t$ to an instantaneous motion intensity $e(t)\in\mathbb{R}^{N\times1}$.
Both signals are computed without gradient propagation and cached as per-Gaussian attributes for efficient reuse during training.


\begin{figure}[t]
\centering
\includegraphics[width=\linewidth]{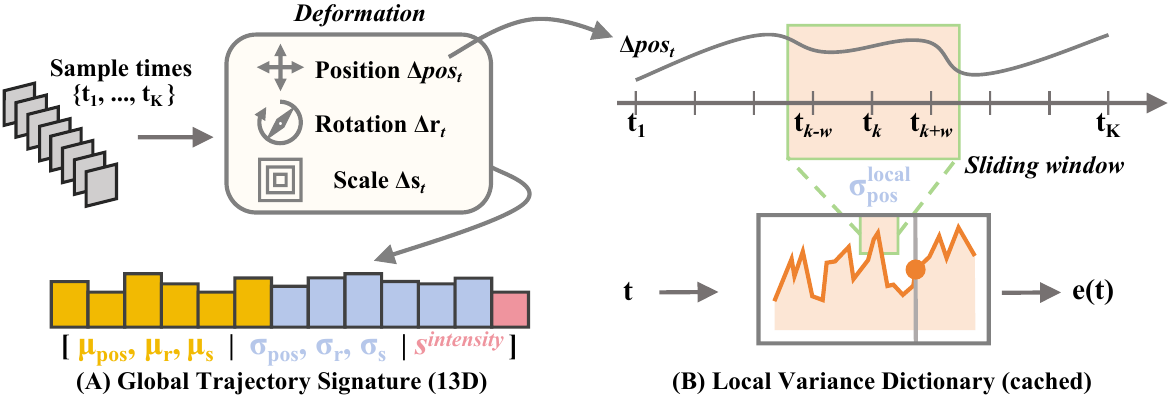}
\caption{
Motion descriptor extraction in MVG.
Given sampled timestamps $\{t_k\}_{k=1}^{K}$, we evaluate each Gaussian's deformation trajectory and compute cross-frame variations in position, rotation, and scale.
(A) The full trajectory is aggregated into a 13D global trajectory signature that captures long-term motion statistics.
(B) Local motion variance is cached as a time-indexed dictionary: given a query timestamp $t$, the dictionary maps it to $e(t)$ by retrieving or interpolating the local variance computed within a sliding temporal window.
}
\label{fig:motion_stats}
\end{figure}

\noindent\textbf{Trajectory Sampling.}
\label{sec:mvg_sampling}
Fig.~\ref{fig:motion_stats} summarizes how we extract the global trajectory signature and the local variance dictionary from simulated deformation trajectories. Every $E$ training iterations, we sample $K$ timestamps $\{t_k\}_{k=1}^{K}$ from the training cameras and evaluate the deformation network for all $N$ dynamic Gaussians:

\begin{equation}
\Delta \mathcal{G}_i(t_k)
=
\bigl(
\Delta\mathbf{pos}_{i,k},\,
\Delta\mathbf{r}_{i,k},\,
\Delta\mathbf{s}_{i,k},\,
\ldots
\bigr),
\quad k=1,\ldots,K .
\end{equation}

We then derive trajectory variations from the predicted deformation residuals. 
In particular, we use the position displacement $\Delta\mathbf{pos}_{i,k}$, 
a scalar rotation magnitude $\theta_{i,k}=\|\Delta\mathbf{r}_{i,k}\|_2$, 
and the log-scale variation $\Delta\bm{\ell}_{i,k}=\log\!\bigl(\mathbf{s}_{i}+\Delta\mathbf{s}_{i,k}\bigr)-\log(\mathbf{s}_{i})$.
The log-scale variation $\Delta\bm{\ell}_{i,k}$ is further decomposed into isotropic and anisotropic components, 
which are later summarized as $\ell^{\text{iso}}$ and $\ell^{\text{aniso}}$ in the global trajectory signature.

\noindent\textbf{Global Motion Trajectory Signature.}
\label{sec:mvg_global}
We summarize the sampled trajectories into a compact global trajectory signature that captures the overall motion behavior of each Gaussian. Specifically, we compute the mean and variance of position, rotation, and scale variations, forming a 13-dimensional signature:

\begin{equation}
\mathbf{v}_i =
\Bigl[
\underbrace{\bar{\Delta\mathbf{pos}}_i,\,
\sigma_{\Delta\mathbf{pos}_i}}_{\text{position (6d)}},
\;
\underbrace{\bar{\theta}_i,\,
\sigma_{\theta_i}}_{\text{rotation (2d)}},
\;
\underbrace{\bar{\ell}^{\text{iso}}_i,\,
\sigma^{\text{iso}}_i}_{\text{scale-iso (2d)}},
\;
\underbrace{\bar{\ell}^{\text{aniso}}_i,\,
\sigma^{\text{aniso}}_i}_{\text{scale-aniso (2d)}},
\;
\underbrace{s_i^{\text{intensity}}}_{\text{motion intensity (1d)}}
\Bigr]^\top ,
\label{eq:mvg_signature}
\end{equation}
where $\bar{\cdot}$ and $\sigma_{\cdot}$ denote temporal mean and variance over the trajectory; the rotation term is 2D because we summarize the scalar rotation magnitude $\theta_{i,k}=\|\Delta\mathbf{r}_{i,k}\|_2$ by its mean and variance.  
The isotropic component $\ell^{\text{iso}}$ measures global scale change, while $\ell^{\text{aniso}}$ captures anisotropic deformation, yielding a more stable representation than directly modeling per-axis scale statistics.
To further summarize motion magnitude, we introduce a scalar motion-intensity score:

\begin{equation}
s_i^{\text{intensity}}
=
0.7\,\sigma_{\text{pos}}(i)
+
0.2\,\sigma_{\text{scale}}(i)
+
0.1\,\sigma_{\text{rot}}(i).
\label{eq:mvg_violence}
\end{equation}

\begin{figure}[t]
\centering
\includegraphics[width=\linewidth]{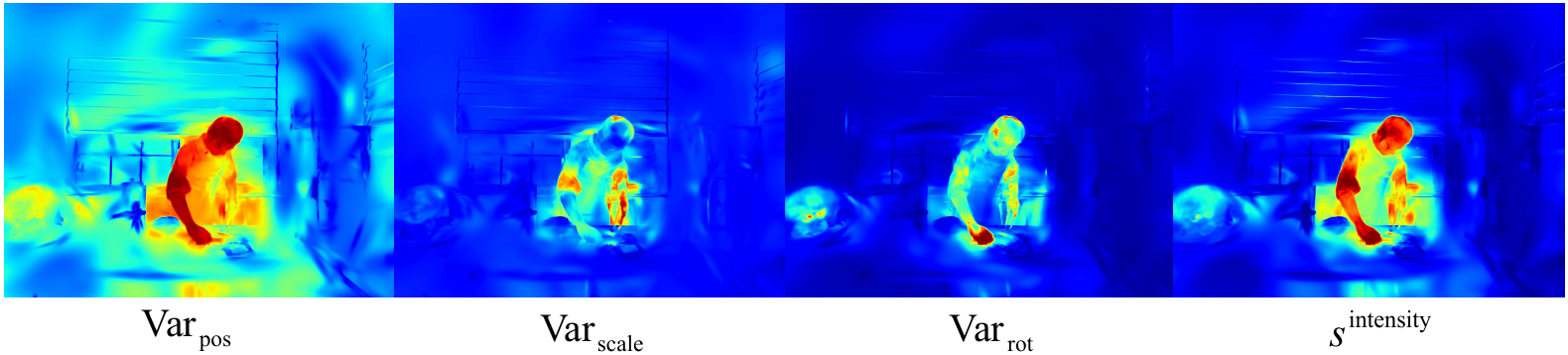}
\caption{
Visualization of per-Gaussian motion variance heatmaps for individual attributes and the fused motion intensity score.
}
\label{fig:var}
\vspace{-0.3cm}
\end{figure}

This score combines position, scale, and rotation variances with empirically chosen weights. Position deviation captures large displacements, scale variance reflects depth-dependent size changes, and rotation variance indicates local articulated motion. The fused intensity score integrates these complementary cues into a unified motion descriptor with clearer foreground--background contrast, serving as an effective motion prior for deformation feature extraction. Fig.~\ref{fig:var} visualizes the attribute-wise variance maps and the fused score, showing its clearer foreground--background separation.

\begin{figure}[t]
\centering
\includegraphics[scale=0.5]{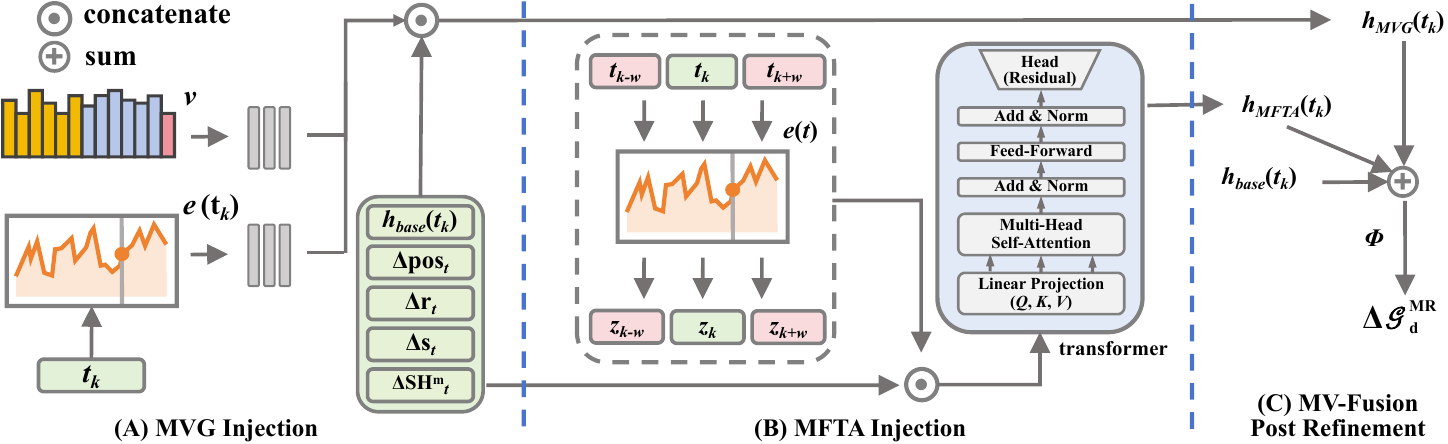}
\caption{
Feature-space motion injection and fusion.
(A) MVG encodes the cached global signature $\mathbf v_i$ and local intensity $e_i(t)$ into a residual feature $h_{\text{MVG}}$.
(B) MFTA forms temporal tokens $\mathbf z_{i,t}$ over a short window and aggregates them via cross-attention to obtain $h_{\text{MFTA}}$.
(C) The fused feature is decoded by the unchanged deformation heads $\Phi$ to predict refined deformation $\Delta \mathcal{G}_d^{\text{MR}}$.
}
\label{fig:injection}
\vspace{-0.5cm}
\end{figure}

\noindent\textbf{Local Variance Dictionary.}
\label{sec:mvg_local}
While the global signature $\mathbf{v}_i$ captures long-term motion statistics, it does not describe short-term variations. We therefore introduce a lightweight local variance dictionary to provide instantaneous motion intensity at arbitrary query times. The dictionary stores local variance values at sparsely sampled timestamps and supports interpolation for unseen timestamps.

Given the displacement trajectory $\Delta\mathbf{pos}_{i,k}$, we define a local motion variance

\begin{equation}
e_i(t_k)
=
\sigma
\left(
\{\Delta\mathbf{pos}_{i,j}\}_{j=k-w}^{k+w}
\right),
\label{eq:mvg_energy}
\end{equation}
which measures the local motion variance within a short temporal window around time $t_k$. To reduce storage overhead, the local variance values are stored only at sparsely sampled timestamps. The value $e_i(t)$ at an arbitrary time is obtained by linear interpolation. For example, in a 300-frame sequence, storing only 10 samples per Gaussian reduces storage from 300 to 10 values (about $30\times$ compression) while preserving the temporal structure of the motion signal.

\noindent\textbf{MVG Injection.}
Fig.~\ref{fig:injection} (A) illustrates the MVG feature injection. The cached global trajectory signature and queried local motion intensity are encoded and injected into the deformation feature space. Given a Gaussian queried at time $t$, the deformation network produces a hidden feature $h_{i,t}$ and a coarse deformation prediction $\Phi(h_{i,t})$. We map the global trajectory signature $\mathbf{v}_i$ together with the local motion intensity $e_i(t)$ to the deformation feature dimension using a lightweight MLP, producing a motion feature $\mathbf{h}_{\text{MVG}}\in\mathbb{R}^{N\times H}$. This feature is then fused with the baseline deformation feature before the refinement heads.

Intuitively, MVG introduces trajectory-level motion priors into the deformation network, helping it distinguish static Gaussians from true movers and better model subtle or transient motion.

\subsection{MotionFormer Temporal Attention (MFTA)}
\label{sec:mfta}

\noindent\textbf{Motivation.}
While MVG provides trajectory-level motion statistics, it does not explicitly capture short-term temporal interactions around the current frame. Such local temporal context is crucial for resolving ambiguous motion patterns. We therefore introduce a lightweight temporal attention module, termed MotionFormer Temporal Attention (MFTA), to aggregate motion cues within a short temporal window as shown in Fig.~\ref{fig:injection} (B).

\noindent\textbf{Temporal Feature Construction.}
Inspired by Timeformer ~\cite{timeformer}, for a Gaussian queried at time $t_i$, we construct a temporal neighborhood $\mathcal{T}(t_i)=\{t_{i-w},t_i,t_{i+w}\}$. For each timestamp $t\in\mathcal{T}(t_i)$, we use the corresponding deformation feature $h_{i,t}$, the predicted deformation value $\Phi(h_{i,t})$, and the local motion intensity $e_i(t)$ to form temporal tokens

\begin{equation}
\mathbf{z}_{i,t} = [\,h_{i,t},\, \Phi(h_{i,t}),\, e_i(t)\,].
\end{equation}

Here, $e_i(t)$ provides an instantaneous cue of motion intensity, helping the attention module emphasize informative temporal neighbors.

\noindent\textbf{MotionFormer Attention.}
To aggregate temporal information, we perform cross-attention ~\cite{transformer} where the current token attends to its temporal neighbors:
\begin{equation}
\mathbf{a}_{i,t_i}
=
\mathrm{Attn}\!\left(
\mathbf{z}_{i,t_i},
\mathbf{z}_{i,t_{i-w}},
\mathbf{z}_{i,t_{i+w}}
\right),
\end{equation}
where $\mathrm{Attn}(Q,K,V)$ denotes a standard cross-attention operator with $Q=\mathbf{z}_{i,t_i}$.

\noindent\textbf{MFTA Injection.}
The attention output is mapped by a lightweight MLP to produce a temporal motion feature $\mathbf{h}_{\text{MFTA}}\in\mathbb{R}^{N\times H}$, which is fused with the baseline deformation feature to form the final motion-aware representation.
\subsection{Training Strategy}
\label{sec:training}

Training proceeds in three stages with progressively activated deformation components.

\noindent\textbf{Stage 1: Base Geometry.}
In the first stage, only the Gaussian geometry is optimized while the deformation network is disabled. 
Aggressive densification and opacity reset are used to establish the initial scene structure.

\noindent\textbf{Stage 2: Base Deformation.}
Once the geometry stabilizes, the baseline deformation network is activated to learn the fundamental spatiotemporal transformations. Near the end of this stage, the temporal branch is enabled to begin collecting motion statistics.

\noindent\textbf{Stage 3: Motion-Aware Refinement.}
In the final stage, the full motion-aware refinement is activated. The zero-initialized MVG and MFTA branches are fused with the baseline deformation feature and trained end-to-end to produce refined deformation updates.

\noindent\textbf{Inference.}
During training, the MVG motion descriptors are computed without back-propagation and cached for reuse. At inference time, the cached MVG descriptors are directly reused, and the temporal module can optionally be disabled for additional efficiency.

\noindent\textbf{Loss Function.}
The training objective combines photometric and regularization losses:
\begin{equation}
\mathcal{L}
=
\lambda_1 \mathcal{L}_{\text{rgb}}
+
\lambda_2 \mathcal{L}_{\text{ssim}}
+
\lambda_3 \mathcal{L}_{\text{reg}},
\end{equation}
Here $\mathcal{L}_{\text{rgb}}$ is the L1 photometric loss, $\mathcal{L}_{\text{ssim}}$ encourages structural similarity, and $\mathcal{L}_{\text{reg}}$ includes standard Gaussian regularization terms (e.g., scale and aspect-ratio constraints).

\noindent\textbf{Implementation Details.}
Our implementation is built upon the DeGauss framework in PyTorch. Training runs for 30k iterations using Adam on a single NVIDIA RTX 3090 GPU. Motion statistics are updated every 2000 iterations by sampling $K{=}64$ timestamps to estimate trajectory variance. For the temporal module, a short temporal window of $w{=}5$ neighboring frames is used to aggregate temporal context. All other settings follow the default configuration of the DeGauss framework.

\section{Experiments}
\subsection{Experimental Settings}
We evaluate our method on dynamic scene reconstruction and distractor-free reconstruction benchmarks. Additional implementation details and sensitivity analyses are provided in the supplementary material.

\textbf{NeRF On-the-Go}~\cite{nerfonthego} contains casually captured real-world scenes with transient distractors such as pedestrians, cyclists, and vehicles. Each scene contains roughly one hundred images with varying occlusion levels, along with a small set of clean hold-out views for evaluation. Following DeGauss, we use seven scenes and train on the occluded images while evaluating novel view synthesis on the clean views.

\textbf{RobustNeRF}~\cite{sabour2023robustnerf} contains static scenes with manually placed distractors that are repositioned or removed across captures, providing a complementary benchmark for distractor-free reconstruction under cross-view inconsistencies.

\textbf{Neu3D}~\cite{neu3d} is a multi-view dynamic video dataset captured by around 20 synchronized static cameras over approximately 300 frames. Following the standard protocol, camera view 0 is used for testing and the remaining views for training.
We report PSNR, SSIM, and LPIPS for image quality assessment.

\subsection {Distractor-Free Scene Reconstruction}

\vspace{-0.5cm}

\begin{table*}[h]
\centering
\caption{Distractor-free scene reconstruction on NeRF On-the-go Dataset\cite{nerfonthego}. The \textcolor{red}{best}, \textcolor{orange}{second best}, and \textcolor{yellow}{third best} are highlighted. Our method shows generally superior performance over state-of-the-art methods.}
\label{tab:table1}
\resizebox{\textwidth}{!}{
\begin{tabular}{l *{21}{c}}
\toprule
 & \multicolumn{3}{c}{Mountain} & \multicolumn{3}{c}{Fountain} & \multicolumn{3}{c}{Corner} & \multicolumn{3}{c}{Patio} & \multicolumn{3}{c}{Spot} & \multicolumn{3}{c}{Patio-High} & \multicolumn{3}{c}{Mean} \\
\cmidrule(lr){2-4} \cmidrule(lr){5-7} \cmidrule(lr){8-10} \cmidrule(lr){11-13} \cmidrule(lr){14-16} \cmidrule(lr){17-19} \cmidrule(lr){20-22}
 & PSNR$\uparrow$ & SSIM$\uparrow$ & LPIPS$\downarrow$ & PSNR$\uparrow$ & SSIM$\uparrow$ & LPIPS$\downarrow$ & PSNR$\uparrow$ & SSIM$\uparrow$ & LPIPS$\downarrow$ & PSNR$\uparrow$ & SSIM$\uparrow$ & LPIPS$\downarrow$ & PSNR$\uparrow$ & SSIM$\uparrow$ & LPIPS$\downarrow$ & PSNR$\uparrow$ & SSIM$\uparrow$ & LPIPS$\downarrow$ & PSNR$\uparrow$ & SSIM$\uparrow$ & LPIPS$\downarrow$ \\
\midrule
RobustNeRF \cite{sabour2023robustnerf} & 17.54 & 0.496 & 0.383 & 15.65 & 0.318 & 0.576 & 23.04 & 0.764 & 0.244 & 20.39 & 0.718 & 0.251 & 20.65 & 0.625 & 0.391 & 20.54 & 0.578 & 0.366 & 19.64 & 0.583 & 0.369 \\
NeRF On-the-go \cite{nerfonthego} & 20.15 & 0.644 & 0.259 & 20.11 & 0.609 & 0.314 & 24.22 & 0.806 & 0.190 & 20.78 & 0.754 & 0.219 & 23.33 & 0.787 & 0.189 & 21.41 & 0.718 & 0.235 & 21.67 & 0.720 & 0.234 \\
3DGS \cite{3dgs} & 19.40 & 0.638 & 0.213 & 19.96 & 0.659 & 0.185 & 20.90 & 0.713 & 0.241 & 17.48 & 0.704 & 0.199 & 20.77 & 0.693 & 0.316 & 17.29 & 0.604 & 0.363 & 19.30 & 0.668 & 0.253 \\
WildGaussians \cite{wildgaussians} & 20.43 & 0.653 & 0.255 & 20.81 & 0.662 & 0.215 & 24.16 & 0.822 & 0.139 & 21.44 & 0.800 & 0.138 & 23.82 & 0.816 & 0.138 & 22.23 & 0.725 & 0.206 & 22.16 & 0.746 & 0.182 \\
DeSplat \cite{Wang2024DeSplatDG} & 19.59 & 0.715 & \cellcolor{yellow!40}0.175 & 20.27 & 0.685 & 0.175 & \cellcolor{red!40}26.05 & \cellcolor{red!40}0.885 & \cellcolor{yellow!40}0.095 & 20.89 & 0.815 & 0.115 & \cellcolor{yellow!40}26.07 & \cellcolor{red!40}0.905 & \cellcolor{yellow!40}0.095 & 22.59 & \cellcolor{red!40}0.845 & \cellcolor{yellow!40}0.125 & 22.58 & \cellcolor{yellow!40}0.813 & \cellcolor{yellow!40}0.130 \\
SpotlessSplats \cite{spotlesssplats} & \cellcolor{yellow!40}21.64 & \cellcolor{yellow!40}0.725 & 0.195 & \cellcolor{orange!40}22.38 & \cellcolor{red!40}0.768 & \cellcolor{yellow!40}0.166 & \cellcolor{yellow!40}25.77 & \cellcolor{orange!40}0.877 & 0.117 & \cellcolor{yellow!40}22.40 & \cellcolor{yellow!40}0.833 & \cellcolor{yellow!40}0.108 & 25.35 & 0.866 & 0.127 & \cellcolor{yellow!40}22.98 & \cellcolor{orange!40}0.808 & 0.155 & \cellcolor{yellow!40}23.42 & \cellcolor{yellow!40}0.813 & 0.145 \\
DeGauss \cite{degauss} & \cellcolor{orange!40}22.31 & \cellcolor{orange!40}0.746 & \cellcolor{orange!40}0.163 & \cellcolor{red!40}22.40 & \cellcolor{orange!40}0.764 & \cellcolor{orange!40}0.139 & \cellcolor{orange!40}25.94 & \cellcolor{yellow!40}0.869 & \cellcolor{orange!40}0.078 & \cellcolor{red!40}22.88 & \cellcolor{orange!40}0.850 & \cellcolor{orange!40}0.087 & \cellcolor{orange!40}26.59 & \cellcolor{yellow!40}0.886 & \cellcolor{orange!40}0.089 & \cellcolor{red!40}23.35 & 0.799 & \cellcolor{orange!40}0.124 & \cellcolor{orange!40}23.91 & \cellcolor{orange!40}0.819 & \cellcolor{orange!40}0.113 \\
Ours & \cellcolor{red!40}22.44 & \cellcolor{red!40}0.755 & \cellcolor{red!40}0.127 & \cellcolor{yellow!40}22.33 & \cellcolor{yellow!40}0.764 & \cellcolor{red!40}0.119 & 25.58 & 0.868 & \cellcolor{red!40}0.061 & \cellcolor{orange!40}22.77 & \cellcolor{red!40}0.870 & \cellcolor{red!40}0.047 & \cellcolor{red!40}27.31 & \cellcolor{orange!40}0.891 & \cellcolor{red!40}0.057 & \cellcolor{orange!40}23.20 & \cellcolor{yellow!40}0.805 & \cellcolor{red!40}0.097 & \cellcolor{red!40}23.94 & \cellcolor{red!40}0.826 & \cellcolor{red!40}0.085 \\
\bottomrule
\vspace{-1cm}
\end{tabular}
}
\end{table*}

\begin{table*}[!t]
\centering
\caption{Quantitative comparison of foreground dynamic region reconstruction on the NeRF On-the-go benchmark. Our method outperforms DeGauss across all scenes and all metrics.}
\label{tab:table2}
\resizebox{\textwidth}{!}{
\begin{tabular}{l *{21}{c}}
\toprule
 & \multicolumn{3}{c}{Mountain} & \multicolumn{3}{c}{Fountain} & \multicolumn{3}{c}{Corner} & \multicolumn{3}{c}{Patio} & \multicolumn{3}{c}{Spot} & \multicolumn{3}{c}{Patio-High} & \multicolumn{3}{c}{Mean} \\
\cmidrule(lr){2-4} \cmidrule(lr){5-7} \cmidrule(lr){8-10} \cmidrule(lr){11-13} \cmidrule(lr){14-16} \cmidrule(lr){17-19} \cmidrule(lr){20-22}
 & PSNR$\uparrow$ & SSIM$\uparrow$ & LPIPS$\downarrow$ & PSNR$\uparrow$ & SSIM$\uparrow$ & LPIPS$\downarrow$ & PSNR$\uparrow$ & SSIM$\uparrow$ & LPIPS$\downarrow$ & PSNR$\uparrow$ & SSIM$\uparrow$ & LPIPS$\downarrow$ & PSNR$\uparrow$ & SSIM$\uparrow$ & LPIPS$\downarrow$ & PSNR$\uparrow$ & SSIM$\uparrow$ & LPIPS$\downarrow$ & PSNR$\uparrow$ & SSIM$\uparrow$ & LPIPS$\downarrow$ \\
\midrule
DeGauss & 21.81 & 0.803 & 0.126 & 19.79 & 0.694 & 0.175 & 21.70 & 0.851 & 0.123 & 25.56 & 0.903 & 0.076 & 27.11 & 0.832 & 0.189 & 21.63 & 0.689 & 0.323 & 22.93 & 0.795 & 0.169 \\
Ours & \textbf{25.62} & \textbf{0.851} & \textbf{0.092} & \textbf{22.52} & \textbf{0.808} & \textbf{0.091} & \textbf{28.23} & \textbf{0.919} & \textbf{0.053} & \textbf{31.57} & \textbf{0.941} & \textbf{0.039} & \textbf{28.92} & \textbf{0.880} & \textbf{0.129} & \textbf{28.73} & \textbf{0.869} & \textbf{0.149} & \textbf{27.60} & \textbf{0.878} & \textbf{0.092} \\
\bottomrule
\end{tabular}
}
\end{table*}

\begin{figure*}[t]
\centering
    \includegraphics[width=0.8\linewidth]{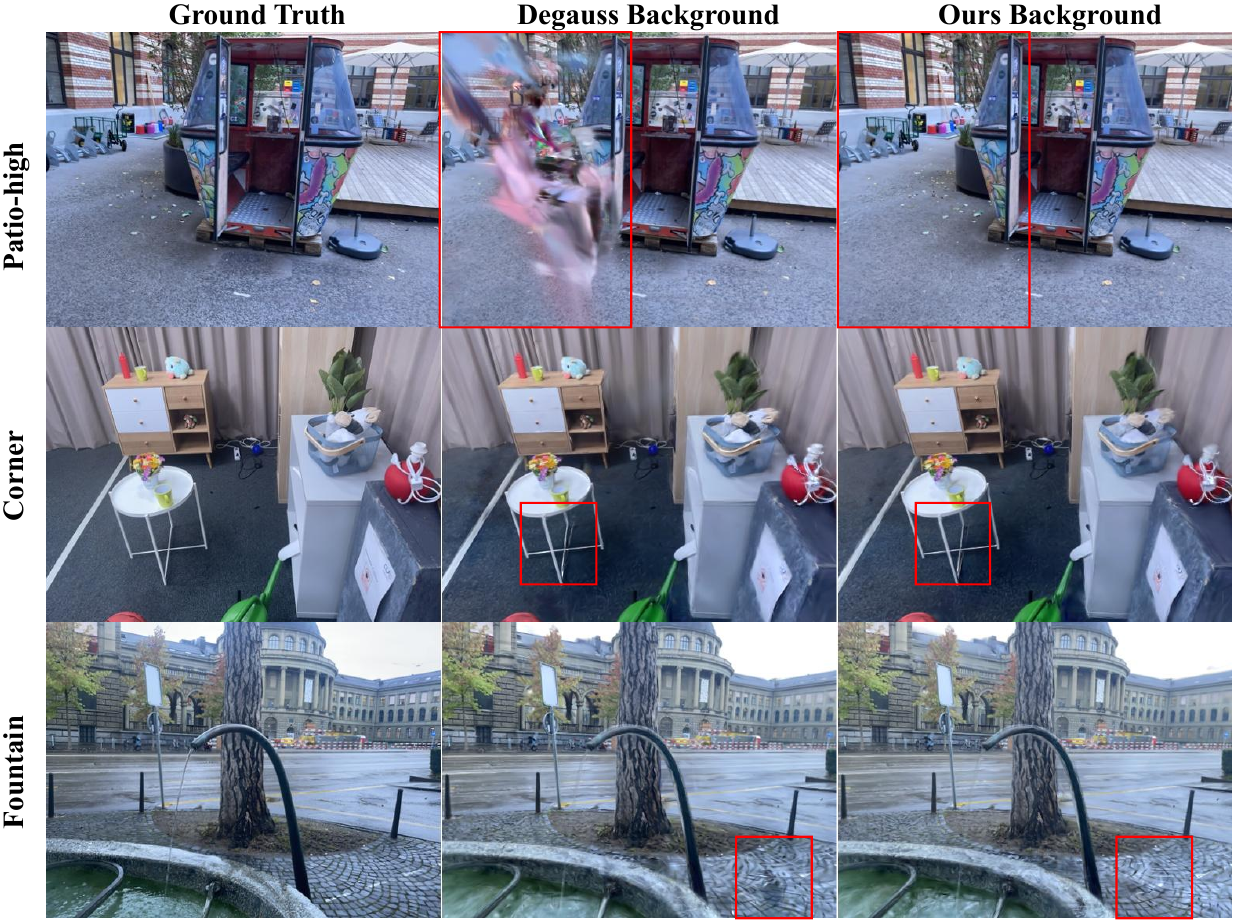}
    \caption{Qualitative comparison of static background reconstruction on held-out test views from the NeRF On-the-go benchmark. Red boxes highlight regions where DeGauss retains visible pseudo-static foreground residuals (e.g., ghostly pedestrian silhouettes in Patio-High), while our method produces substantially cleaner backgrounds.}
    \label{fig:onthego_static}
    \vspace{-0.6cm}
\end{figure*}

\begin{figure*}[t]
    \centering
    \includegraphics[width=0.92\linewidth]{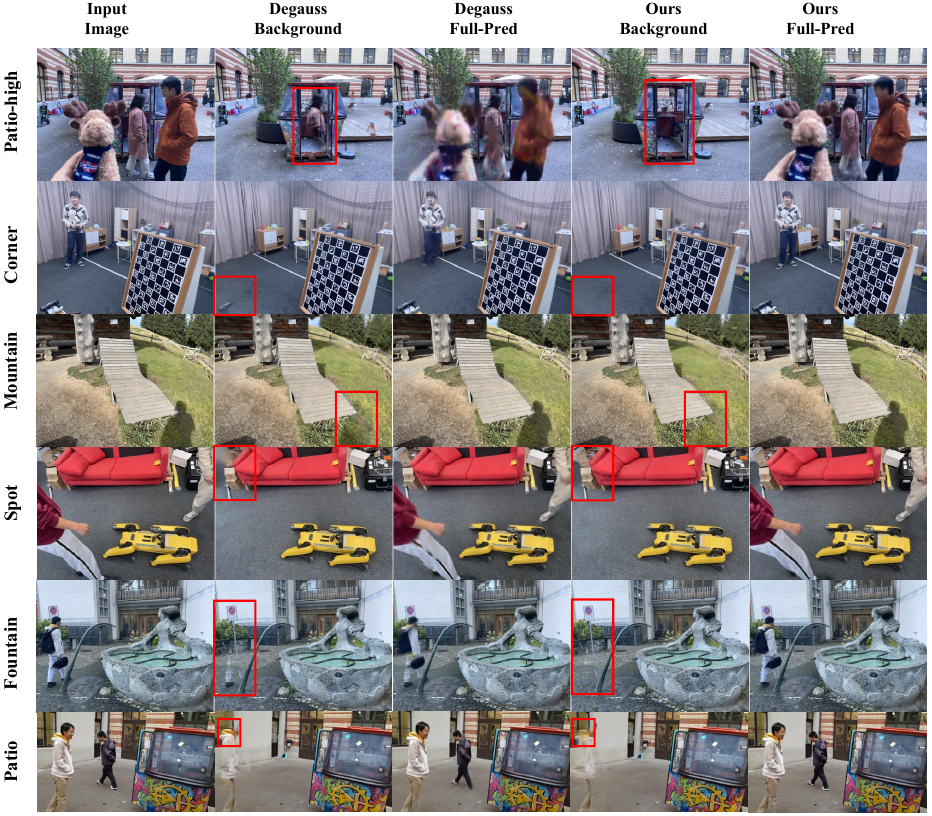}
    \caption{Qualitative comparison of background-only and full composed renderings on NeRF On-the-go benchmark. Red boxes reveal that DeGauss backgrounds retain ghostly foreground residuals, while our method achieves cleaner backgrounds and sharper full-scene compositions.}
    \label{fig:onthego_foreground}
    \vspace{-0.3cm}
\end{figure*}

Table~\ref{tab:table1} presents the quantitative comparison on the NeRF On-the-go benchmark for distractor-free static scene reconstruction. Our method ranks first across all three averaged metrics and achieves the lowest LPIPS on all six scenes, indicating consistently superior perceptual quality. Even on scenes where our PSNR is marginally below DeGauss, we obtain substantially better LPIPS, confirming that enhanced motion awareness leads to cleaner and more artifact-free background reconstruction.

We further report quantitative results for dynamic foreground regions in Table~\ref{tab:table2}. Our method consistently outperforms DeGauss across all scenes and metrics, demonstrating that the proposed motion-aware refinement not only improves background stability but also enhances the reconstruction of dynamic objects.

\begin{figure*}[!t]
    \centering
    \includegraphics[width=0.8\linewidth]{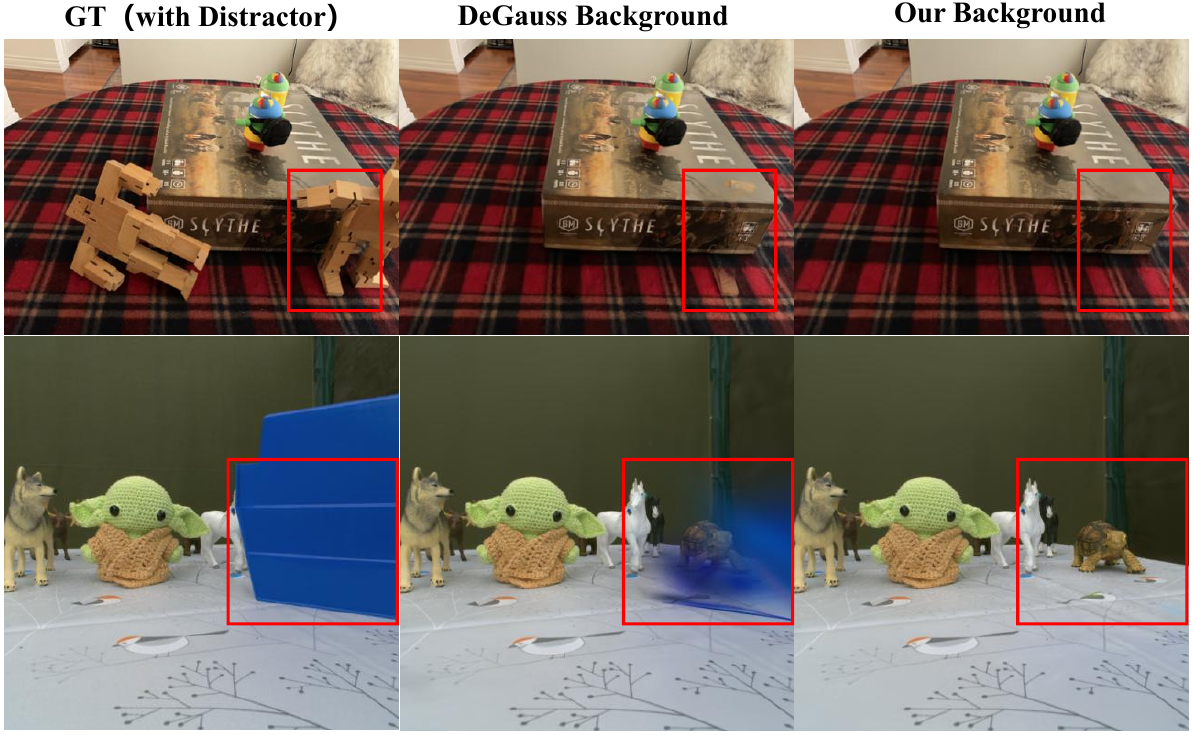}
    \caption{Qualitative comparison on RobustNeRF. MVFusion-GS produces cleaner backgrounds than DeGauss.}
    \label{fig:robustnerf}
     \vspace{-0.8cm}
\end{figure*}

Fig.s ~\ref{fig:onthego_static} and ~\ref{fig:onthego_foreground} visualize the background reconstruction quality on the NeRF On-the-Go benchmark. As highlighted by the red boxes, DeGauss retains pseudo-static foreground residuals in the background branch that are not captured by the deformation network, including a blurred pedestrian silhouette in Patio-High and diffuse smearing in Corner and Fountain. In contrast, MVFusion-GS substantially removes these artifacts, producing cleaner backgrounds with sharper edges and textures. The full composed predictions in Fig.~\ref{fig:onthego_foreground} further show that the improved dynamic-static separation preserves overall rendering quality, while the held-out test views in Fig.~\ref{fig:onthego_static} demonstrate generalization beyond the training views.

\begin{table*}[h]
\centering
\caption{Quantitative comparison on the RobustNeRF~\cite{sabour2023robustnerf} benchmark. The \textcolor{red}{best}, \textcolor{orange}{second best}, and \textcolor{yellow}{third best} results are highlighted.}
\label{tab:robustnerf}
\resizebox{\textwidth}{!}{
\begin{tabular}{l *{15}{c}}
\toprule
 & \multicolumn{3}{c}{Android} & \multicolumn{3}{c}{Crab2} & \multicolumn{3}{c}{Statue} & \multicolumn{3}{c}{Yoda} & \multicolumn{3}{c}{Mean} \\
\cmidrule(lr){2-4} \cmidrule(lr){5-7} \cmidrule(lr){8-10} \cmidrule(lr){11-13} \cmidrule(lr){14-16}
 & PSNR$\uparrow$ & SSIM$\uparrow$ & LPIPS$\downarrow$ & PSNR$\uparrow$ & SSIM$\uparrow$ & LPIPS$\downarrow$ & PSNR$\uparrow$ & SSIM$\uparrow$ & LPIPS$\downarrow$ & PSNR$\uparrow$ & SSIM$\uparrow$ & LPIPS$\downarrow$ & PSNR$\uparrow$ & SSIM$\uparrow$ & LPIPS$\downarrow$ \\
\midrule
3DGS~\cite{3dgs} & 23.32 & 0.794 & 0.159 & 31.76 & 0.925 & 0.172 & 20.83 & 0.830 & 0.148 & 28.92 & 0.905 & 0.192 & 26.21 & 0.863 & 0.168 \\
WildGaussians~\cite{wildgaussians} & \cellcolor{red!40}24.67 & \cellcolor{red!40}0.828 & \cellcolor{yellow!40}0.151 & 30.52 & 0.909 & 0.213 & \cellcolor{yellow!40}22.54 & \cellcolor{red!40}0.863 & \cellcolor{yellow!40}0.129 & 30.55 & 0.905 & 0.202 & 27.07 & \cellcolor{yellow!40}0.876 & 0.174 \\
SpotlessSplats~\cite{spotlesssplats} & 24.20 & 0.810 & 0.159 & \cellcolor{yellow!40}33.90 & \cellcolor{yellow!40}0.933 & \cellcolor{yellow!40}0.169 & 21.97 & 0.821 & 0.163 & \cellcolor{orange!40}34.24 & \cellcolor{yellow!40}0.938 & \cellcolor{yellow!40}0.156 & \cellcolor{yellow!40}28.58 & 0.875 & \cellcolor{yellow!40}0.162 \\
DeGauss~\cite{degauss} & \cellcolor{yellow!40}24.54 & \cellcolor{yellow!40}0.813 & \cellcolor{orange!40}0.083 & \cellcolor{orange!40}34.48 & \cellcolor{orange!40}0.952 & \cellcolor{orange!40}0.076 & \cellcolor{red!40}23.08 & \cellcolor{yellow!40}0.861 & \cellcolor{orange!40}0.097 & \cellcolor{yellow!40}33.48 & \cellcolor{orange!40}0.947 & \cellcolor{orange!40}0.082 & \cellcolor{orange!40}28.89 & \cellcolor{orange!40}0.893 & \cellcolor{orange!40}0.085 \\
MVFusion-GS (Ours) & \cellcolor{orange!40}24.58 & \cellcolor{orange!40}0.814 & \cellcolor{red!40}0.068 & \cellcolor{red!40}34.87 & \cellcolor{red!40}0.955 & \cellcolor{red!40}0.065 & \cellcolor{orange!40}23.07 & \cellcolor{orange!40}0.863 & \cellcolor{red!40}0.070 & \cellcolor{red!40}35.03 & \cellcolor{red!40}0.955 & \cellcolor{red!40}0.074 & \cellcolor{red!40}29.39 & \cellcolor{red!40}0.897 & \cellcolor{red!40}0.069 \\
\bottomrule
\end{tabular}
}
\end{table*}

Table~\ref{tab:robustnerf} and Fig.~\ref{fig:robustnerf} further show that MVFusion-GS generalizes to RobustNeRF, where manually repositioned distractors create cross-view inconsistencies rather than continuous motion. The gains indicate that our motion-aware refinement also handles distractors that appear or disappear across views.

\subsection{Dynamic Scene Reconstruction Results}

\vspace{-0.2cm}

\begin{table*}[h]
\centering
\caption{Quantitative comparison on the Neu3D dataset. \textit{MVFusion-GS (4DGS plug-in)} applies the proposed MVG/MFTA refinement to the single-branch 4DGS deformation pipeline, while \textit{MVFusion-GS (Ours)} denotes the full decoupled model.}
\label{tab:dynerf}
\resizebox{\textwidth}{!}{
\begin{tabular}{l *{21}{c}}
\toprule
 & \multicolumn{3}{c}{Cut Beef} & \multicolumn{3}{c}{Cook Spinach} & \multicolumn{3}{c}{Sear Steak} & \multicolumn{3}{c}{Flame Steak} & \multicolumn{3}{c}{Flame Salmon} & \multicolumn{3}{c}{Coffee Martini} & \multicolumn{3}{c}{Mean} \\
\cmidrule(lr){2-4} \cmidrule(lr){5-7} \cmidrule(lr){8-10} \cmidrule(lr){11-13} \cmidrule(lr){14-16} \cmidrule(lr){17-19} \cmidrule(lr){20-22}
 & PSNR$\uparrow$ & SSIM$\uparrow$ & LPIPS$\downarrow$
 & PSNR$\uparrow$ & SSIM$\uparrow$ & LPIPS$\downarrow$
 & PSNR$\uparrow$ & SSIM$\uparrow$ & LPIPS$\downarrow$
 & PSNR$\uparrow$ & SSIM$\uparrow$ & LPIPS$\downarrow$
 & PSNR$\uparrow$ & SSIM$\uparrow$ & LPIPS$\downarrow$
 & PSNR$\uparrow$ & SSIM$\uparrow$ & LPIPS$\downarrow$
 & PSNR$\uparrow$ & SSIM$\uparrow$ & LPIPS$\downarrow$ \\
\midrule
NeRFPlayer~\cite{nerfplayer} & 31.83 & 0.928 & 0.119 & 32.06 & 0.930 & 0.116 & 32.31 & 0.940 & 0.111 & 27.36 & 0.867 & 0.215 & 26.14 & 0.849 & 0.233 & \cellcolor{red!40}32.05 & \cellcolor{yellow!40}0.938 & 0.111 & 30.29 & 0.909 & 0.151 \\
HexPlane~\cite{hexplane} & 30.83 & 0.927 & 0.115 & 31.05 & 0.928 & 0.114 & 30.00 & 0.939 & 0.105 & 30.42 & 0.939 & 0.104 & 29.23 & 0.905 & 0.088 & 28.45 & 0.891 & 0.149 & 30.00 & 0.922 & 0.113 \\
K-Planes~\cite{kplanes} & 31.82 & \cellcolor{red!40}0.966 & 0.114 & 32.60 & \cellcolor{red!40}0.966 & 0.114 & 32.52 & \cellcolor{red!40}0.974 & 0.104 & 32.39 & \cellcolor{red!40}0.970 & 0.102 & \cellcolor{red!40}30.44 & \cellcolor{red!40}0.953 & 0.132 & \cellcolor{orange!40}29.99 & \cellcolor{red!40}0.953 & 0.134 & 31.63 & \cellcolor{red!40}0.964 & 0.117 \\
MixVoxels~\cite{mixvoxels} & 31.30 & \cellcolor{orange!40}0.965 & 0.111 & 31.65 & \cellcolor{orange!40}0.965 & 0.113 & 31.43 & \cellcolor{orange!40}0.971 & 0.103 & 31.21 & \cellcolor{red!40}0.970 & 0.108 & \cellcolor{orange!40}29.92 & \cellcolor{orange!40}0.945 & 0.163 & \cellcolor{yellow!40}29.36 & \cellcolor{orange!40}0.946 & 0.147 & 30.81 & \cellcolor{orange!40}0.960 & 0.124 \\
SWinGS~\cite{swings} & 31.84 & 0.945 & 0.099 & 31.96 & 0.946 & 0.094 & 32.21 & 0.950 & 0.092 & 32.18 & 0.953 & 0.087 & \cellcolor{yellow!40}29.25 & \cellcolor{yellow!40}0.925 & 0.100 & 29.25 & 0.925 & 0.100 & 31.12 & 0.941 & 0.095 \\
4DGS~\cite{4dgs} & 32.66 & 0.946 & 0.053 & 32.46 & 0.949 & 0.052 & 32.49 & 0.957 & 0.041 & 32.75 & 0.954 & 0.040 & 29.00 & 0.912 & 0.081 & 27.34 & 0.905 & 0.083 & 31.12 & 0.937 & 0.058 \\
MangoGS~\cite{mangogs} & \cellcolor{yellow!40}33.28 & 0.949 & \cellcolor{yellow!40}0.043 & \cellcolor{yellow!40}32.83 & 0.947 & \cellcolor{yellow!40}0.042 & \cellcolor{yellow!40}33.09 & 0.953 & 0.043 & \cellcolor{orange!40}34.11 & \cellcolor{yellow!40}0.958 & \cellcolor{yellow!40}0.035 & 29.10 & 0.918 & \cellcolor{yellow!40}0.074 & 28.95 & 0.916 & \cellcolor{yellow!40}0.073 & \cellcolor{orange!40}31.89 & 0.940 & \cellcolor{yellow!40}0.052 \\
DeGauss~\cite{degauss} & 32.56 & 0.957 & \cellcolor{orange!40}0.042 & 32.61 & 0.950 & \cellcolor{orange!40}0.041 & \cellcolor{orange!40}33.20 & 0.956 & \cellcolor{orange!40}0.035 & 32.75 & 0.955 & \cellcolor{orange!40}0.034 & 29.23 & 0.916 & \cellcolor{red!40}0.068 & 28.80 & 0.916 & \cellcolor{red!40}0.062 & 31.52 & 0.942 & \cellcolor{orange!40}0.047 \\
\midrule
MVFusion-GS (4DGS plug-in) & \cellcolor{orange!40}33.31 & 0.952 & 0.052 & \cellcolor{orange!40}33.01 & \cellcolor{yellow!40}0.954 & 0.051 & 33.04 & \cellcolor{yellow!40}0.960 & \cellcolor{yellow!40}0.040 & \cellcolor{yellow!40}33.45 & 0.957 & 0.039 & \cellcolor{yellow!40}29.25 & 0.918 & 0.080 & 27.90 & 0.911 & 0.080 & \cellcolor{yellow!40}31.66 & 0.942 & 0.057 \\
MVFusion-GS (Ours) & \cellcolor{red!40}33.62 & \cellcolor{yellow!40}0.960 & \cellcolor{red!40}0.040 & \cellcolor{red!40}33.41 & \cellcolor{yellow!40}0.954 & \cellcolor{red!40}0.039 & \cellcolor{red!40}33.73 & 0.957 & \cellcolor{red!40}0.033 & \cellcolor{red!40}34.13 & \cellcolor{orange!40}0.959 & \cellcolor{red!40}0.032 & 28.89 & 0.915 & \cellcolor{orange!40}0.070 & 28.63 & 0.916 & \cellcolor{orange!40}0.065 & \cellcolor{red!40}32.07 & \cellcolor{yellow!40}0.943 & \cellcolor{red!40}0.046 \\
\bottomrule
\end{tabular}
}
\end{table*}

\begin{figure*}[t]
    \centering
    \includegraphics[width=1\linewidth]{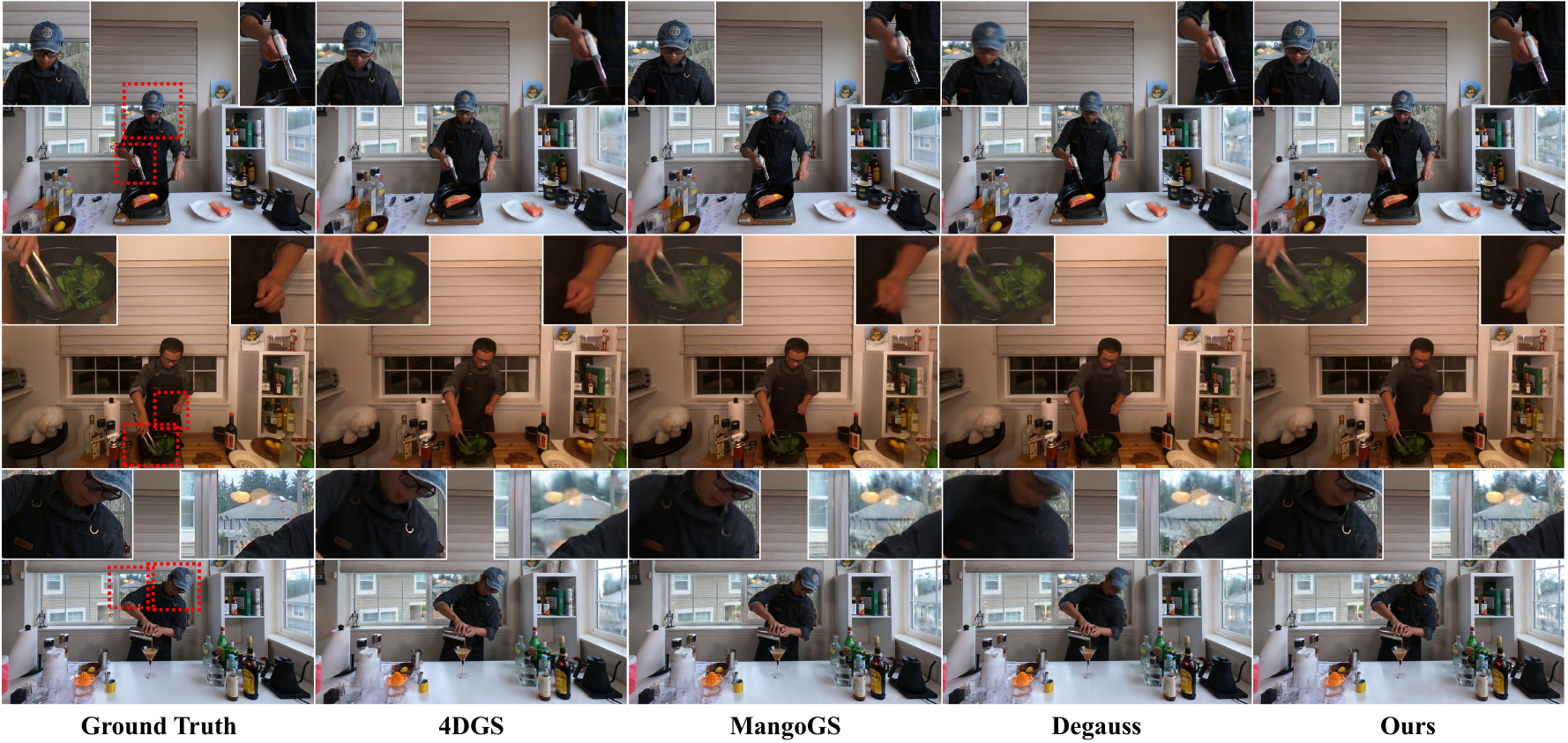}
    \caption{Qualitative comparison on Neu3D. MVFusion-GS reconstructs sharper motion details with fewer artifacts.}
    \label{fig:dynerf_compare}
    \vspace{-0.3cm}
\end{figure*}

\begin{table}[t]                              
\centering
\small                                                                              
\caption{Ablation study of the proposed motion-aware refinement.}
\scalebox{0.70}{
\begin{tabular}{lcccccc}
\toprule
\multirow{2}{*}{Method} & \multicolumn{3}{c}{Neu3D} & \multicolumn{3}{c}{NeRF On-the-go} \\
\cmidrule(lr){2-4} \cmidrule(lr){5-7}
& PSNR $\uparrow$ & SSIM $\uparrow$ & LPIPS $\downarrow$ & PSNR $\uparrow$ & SSIM $\uparrow$ & LPIPS $\downarrow$ \\
\midrule
w/o MVG \& MFTA (Base deformation) & 31.52 & 0.942 & 0.047 & 23.86 & 0.807 & 0.122 \\
w/o MVG (no motion statistics) & 31.64 & 0.937 & 0.052 & 23.87 & 0.810 & 0.114 \\
w/ MVG (position variance only) & 31.93 & 0.943 & 0.046 & 23.92 & 0.815 & 0.096 \\
w/o MFTA & 31.78 & 0.942 & 0.047 & 23.89 & 0.820 & 0.089 \\
w/o Cross-attention (Self-attention) & 32.01 & 0.943 & 0.046 & 23.93 & 0.822 & 0.090 \\
Full Model & \textbf{32.07} & \textbf{0.943} & \textbf{0.046} & \textbf{23.94} & \textbf{0.826} & \textbf{0.085} \\
\bottomrule
\end{tabular}
}
\vspace{-0.5cm}
\label{tab:ablation}
\end{table}

Table~\ref{tab:dynerf} reports Neu3D results. MVFusion-GS achieves the best mean PSNR and LPIPS, and its unified variant improves over the single-branch baseline, validating the plug-in use of MVG/MFTA. MVFusion-GS also obtains the best dynamic-region PSNR of 31.78, outperforming SpaceTimeGS (30.85) and ST-4DGS (31.35).

Fig.~\ref{fig:dynerf_compare} shows qualitative comparisons from the Neu3D dataset. In highly dynamic regions such as flames and rapidly moving objects, competing methods exhibit motion blur or artifacts, while our approach preserves clearer structural details and more consistent motion patterns. Meanwhile, the improvements are not limited to dynamic regions. Our method also produces cleaner backgrounds with fewer residual artifacts, leading to more coherent full-scene renderings.

\subsection{Ablation Study}
\label{sec:ablation}

Table~\ref{tab:ablation} evaluates the contribution of each component. Starting from the base deformation model without MVG and MFTA, adding the refinement branch without trajectory-level MVG statistics improves Neu3D PSNR from 31.52 to 31.64, while using only positional variance further improves it to 31.93. This shows that motion statistics are useful and that the full global trajectory signature is more effective than position-only motion cues.

Removing the temporal module (w/o MFTA) reduces the performance to 31.78 PSNR, indicating that temporal context aggregation helps capture dynamic motion patterns. Replacing the proposed cross-attention with a simpler self-attention design also lowers the performance to 32.01 PSNR, showing that cross-frame attention is more effective than processing temporal tokens without query-centered aggregation.

The full model achieves the best result, demonstrating that combining motion statistics with cross-frame temporal reasoning provides complementary benefits for dynamic scene reconstruction. Table~\ref{tab:ablation} also reports ablation results on NeRF On-the-go. Since this benchmark evaluates static background reconstruction, the PSNR margins between variants are relatively small, as background regions are inherently less sensitive to deformation accuracy. However, the perceptual metrics reveal clear trends. Further ablations on motion-score fusion weights, motion-statistics update interval, temporal window size, and temporal consistency are reported in the supplementary material.

\section{Conclusion}

We introduced MVFusion-GS, a motion-aware refinement framework for dynamic 3D Gaussian Splatting. By combining motion trajectory statistics and temporal attention, the proposed method improves motion modeling and dynamic–static separation. Experiments demonstrate consistent gains in both dynamic scene reconstruction and distractor-free reconstruction. 

Despite these improvements, our method still depends on the baseline deformation field to provide reliable motion cues, and the estimated variance may become less discriminative when the initial deformation underfits complex foreground motion. In such challenging cases, especially under heavy occlusion or extremely weak motion evidence, residual artifacts may remain and motivate future work on more robust motion-aware Gaussian reassignment.

%
%
\clearpage
\bibliographystyle{splncs04}
\bibliography{main}

@String(CVPR  = {IEEE Conf. Comput. Vis. Pattern Recog.})

@String(ICCV  = {Int. Conf. Comput. Vis.})

@String(NeurIPS = {Adv. Neural Inform. Process. Syst.})

@String(CVPR  = {CVPR})

@String(ICCV  = {ICCV})

@String(NeurIPS = {NeurIPS})

@article{nerf,
author = {Mildenhall, Ben and Srinivasan, Pratul P. and Tancik, Matthew and Barron, Jonathan T. and Ramamoorthi, Ravi and Ng, Ren},
title = {NeRF: representing scenes as neural radiance fields for view synthesis},
year = {2021},
issue_date = {January 2022},
publisher = {Association for Computing Machinery},
address = {New York, NY, USA},
volume = {65},
number = {1},
issn = {0001-0782},
doi = {10.1145/3503250},
journal = {Commun. ACM},
month = dec,
pages = {99–106},
numpages = {8}
}

@inproceedings{mobilenerf,
    title     = {{MobileNeRF}: Exploiting the Polygon Rasterization Pipeline for Efficient Neural Field Rendering on Mobile Architectures},
    author    = {Chen, Zhiqin and Funkhouser, Thomas and Hedman, Peter and Tagliasacchi, Andrea},
    booktitle = {Proceedings of the IEEE/CVF Conference on Computer Vision and Pattern Recognition (CVPR)},
    pages     = {16569--16578},
    year      = {2023}
}

@article{instantngp,
    title     = {Instant Neural Graphics Primitives with a Multiresolution Hash Encoding},
    author    = {M\"uller, Thomas and Evans, Alex and Schied, Christoph and Keller, Alexander},
    journal   = {ACM Transactions on Graphics},
    volume    = {41},
    number    = {4},
    pages     = {102:1--102:15},
    year      = {2022},
    doi       = {10.1145/3528223.3530127},
    publisher = {ACM}
}

@inproceedings{deepsdf,
    title     = {{DeepSDF}: Learning Continuous Signed Distance Functions for Shape Representation},
    author    = {Park, Jeong Joon and Florence, Peter and Straub, Julian and Newcombe, Richard and Lovegrove, Steven},
    booktitle = {Proceedings of the IEEE/CVF Conference on Computer Vision and Pattern Recognition (CVPR)},
    pages     = {165--174},
    year      = {2019}
}

@inproceedings{volsdf,
    title     = {Volume Rendering of Neural Implicit Surfaces},
    author    = {Yariv, Lior and Gu, Jiatao and Kasten, Yoni and Lipman, Yaron},
    booktitle = {Advances in Neural Information Processing Systems (NeurIPS)},
    year      = {2021}
}

@article{3dgs,
    title     = {3D Gaussian Splatting for Real-Time Radiance Field Rendering},
    author    = {Kerbl, Bernhard and Kopanas, Georgios and Leimk{\"u}hler, Thomas and Drettakis, George},
    journal   = {ACM Transactions on Graphics},
    volume    = {42},
    number    = {4},
    year      = {2023},
    url       = {https://repo-sam.inria.fr/fungraph/3d-gaussian-splatting/}
}

@inproceedings{scaffoldgs,
  title={Scaffold-gs: Structured 3d gaussians for view-adaptive rendering},
  author={Lu, Tao and Yu, Mulin and Xu, Linning and Xiangli, Yuanbo and Wang, Limin and Lin, Dahua and Dai, Bo},
  booktitle={Proceedings of the IEEE/CVF Conference on Computer Vision and Pattern Recognition},
  pages={20654--20664},
  year={2024}
}

@inproceedings{2dgs,
author = {Huang, Binbin and Yu, Zehao and Chen, Anpei and Geiger, Andreas and Gao, Shenghua},
title = {2D Gaussian Splatting for Geometrically Accurate Radiance Fields},
year = {2024},
isbn = {9798400705250},
publisher = {Association for Computing Machinery},
address = {New York, NY, USA},
doi = {10.1145/3641519.3657428},
booktitle = {ACM SIGGRAPH 2024 Conference Papers},
articleno = {32},
numpages = {11},
location = {Denver, CO, USA},
series = {SIGGRAPH '24}
}

@inproceedings{deformablegs,
    title     = {Deformable 3D Gaussians for High-Fidelity Monocular Dynamic Scene Reconstruction},
    author    = {Yang, Ziyi and Gao, Xinyu and Zhou, Wen and Jiao, Shaohui and Zhang, Yuqing and Jin, Xiaogang},
    booktitle = {Proceedings of the IEEE/CVF Conference on Computer Vision and Pattern Recognition (CVPR)},
    pages     = {20331--20341},
    year      = {2024}
}

@inproceedings{4dgs,
    author    = {Wu, Guanjun and Yi, Taoran and Fang, Jiemin and Xie, Lingxi and Zhang, Xiaopeng and Wei, Wei and Liu, Wenyu and Tian, Qi and Wang, Xinggang},
    title     = {4D Gaussian Splatting for Real-Time Dynamic Scene Rendering},
    booktitle = {Proceedings of the IEEE/CVF Conference on Computer Vision and Pattern Recognition (CVPR)},
    month     = {June},
    year      = {2024},
    pages     = {20310-20320}
}

@inproceedings{scgs,
    title     = {{SC-GS}: Sparse-Controlled Gaussian Splatting for Editable Dynamic Scenes},
    author    = {Huang, Yi-Hua and Sun, Yang-Tian and Yang, Ziyi and Lyu, Xiaoyang and Cao, Yan-Pei and Qi, Xiaojuan},
    booktitle = {Proceedings of the IEEE/CVF Conference on Computer Vision and Pattern Recognition (CVPR)},
    pages     = {4220--4230},
    year      = {2024}
}

@inproceedings{sabour2023robustnerf,
    author    = {Sabour, Sara and Vora, Suhani and Duckworth, Daniel and Krasin, Ivan and Fleet, David J. and Tagliasacchi, Andrea},
    title     = {{RobustNeRF}: Ignoring Distractors With Robust Losses},
    booktitle = {Proceedings of the IEEE/CVF Conference on Computer Vision and Pattern Recognition (CVPR)},
    month     = {June},
    year      = {2023},
    pages     = {20626-20636}
}

@inproceedings{nerfonthego,
    author    = {Ren, Weining and Zhu, Zihan and Sun, Boyang and Chen, Jiaqi and Pollefeys, Marc and Peng, Songyou},
    title     = {{NeRF On-the-go}: Exploiting Uncertainty for Distractor-free NeRFs in the Wild},
    booktitle = {Proceedings of the IEEE/CVF Conference on Computer Vision and Pattern Recognition (CVPR)},
    month     = {June},
    year      = {2024},
    pages     = {8931-8940}
}

@inproceedings{wildgaussians,
  title={{W}ild{G}aussians: {3D} Gaussian Splatting in the Wild},
  author={Kulhanek, Jonas and Peng, Songyou and Kukelova, Zuzana and Pollefeys, Marc and Sattler, Torsten},
  booktitle={Proceedings of the 38th International Conference on Neural Information Processing Systems},
  year={2024}
}

@article{spotlesssplats,
author = {Sabour, Sara and Goli, Lily and Kopanas, George and Matthews, Mark and Lagun, Dmitry and Guibas, Leonidas and Jacobson, Alec and Fleet, David and Tagliasacchi, Andrea},
title = {SpotLessSplats: Ignoring Distractors in 3D Gaussian Splatting},
year = {2025},
issue_date = {April 2025},
publisher = {Association for Computing Machinery},
address = {New York, NY, USA},
volume = {44},
number = {2},
issn = {0730-0301},
doi = {10.1145/3727143},
journal = {ACM Trans. Graph.},
month = apr,
articleno = {17},
numpages = {11}
}

@inproceedings{martinbrualla2021nerfw,
    author = {Martin-Brualla, Ricardo
            and Radwan, Noha
            and Sajjadi, Mehdi S. M.
            and Barron, Jonathan T.
            and Dosovitskiy, Alexey
            and Duckworth, Daniel},
    title = {{NeRF in the Wild: Neural Radiance Fields for
           Unconstrained Photo Collections}},
    booktitle = {CVPR},
    year={2021}
}

@inproceedings{d2nerf,
author = {Wu, Tianhao and Zhong, Fangcheng and Tagliasacchi, Andrea and Cole, Forrester and Oztireli, Cengiz},
title = {{D$^2$NeRF}: self-supervised decoupling of dynamic and static objects from a monocular video},
year = {2022},
isbn = {9781713871088},
publisher = {Curran Associates Inc.},
address = {Red Hook, NY, USA},
booktitle = {Proceedings of the 36th International Conference on Neural Information Processing Systems},
articleno = {2366},
numpages = {14},
location = {New Orleans, LA, USA},
series = {NIPS '22}
}

@InProceedings{degauss,
    author    = {Wang, Rui and Lohmeyer, Quentin and Meboldt, Mirko and Tang, Siyu},
    title     = {{DeGauss}: Dynamic-Static Decomposition with Gaussian Splatting for Distractor-free 3D Reconstruction},
    booktitle = {Proceedings of the IEEE/CVF International Conference on Computer Vision (ICCV)},
    month     = {October},
    year      = {2025},
    pages     = {6294-6303}
}

@InProceedings{Wang2024DeSplatDG,
    author    = {Wang, Yihao and Klasson, Marcus and Turkulainen, Matias and Wang, Shuzhe and Kannala, Juho and Solin, Arno},
    title     = {{DeSplat}: Decomposed Gaussian Splatting for Distractor-Free Rendering},
    booktitle = {Proceedings of the IEEE/CVF Conference on Computer Vision and Pattern Recognition (CVPR)},
    month     = {June},
    year      = {2025},
    pages     = {722-732}
}

@inproceedings{wang2025shapeofmotion,
  title     = {Shape of Motion: 4D Reconstruction from a Single Video},
  author    = {Wang, Qianqian and Ye, Vickie and Gao, Hang and Zeng, Weijia and Austin, Jake and Li, Zhengqi and Kanazawa, Angjoo},
  booktitle   = {International Conference on Computer Vision (ICCV)},
  year      = {2025}
}

@InProceedings{park2025splinegs,
      author    = {Park, Jongmin and Bui, Minh-Quan Viet and Bello, Juan Luis Gonzalez and Moon, Jaeho and Oh, Jihyong and Kim, Munchurl},
      title     = {{SplineGS}: Robust Motion-Adaptive Spline for Real-Time Dynamic 3D Gaussians from Monocular Video},
      booktitle = {Proceedings of the Computer Vision and Pattern Recognition Conference (CVPR)},
      month     = {June},
      year      = {2025},
      pages     = {26866-26875}
  }

@inproceedings{stearns2024marbles,
  title={Dynamic Gaussian Marbles for Novel View Synthesis of Casual Monocular Videos},
  author={Stearns, Colton and Harley, Adam W and Uy, Mikaela and Dubost, Florian and Tombari, Federico and Wetzstein, Gordon and Guibas, Leonidas},
  booktitle={SIGGRAPH Asia 2024 Conference Papers},
  pages={1--11},
  year={2024}
}

@inproceedings{mangogs,
  title={{Mango-GS}: Enhancing Spatio-Temporal Consistency in Dynamic Scenes Reconstruction using Multi-Frame Node-Guided 4D Gaussian Splatting},
  author={Huang, Tingxuan and Zhu, Haowei and Yong, Jun-Hai and Pan, Hao and Wang, Bin},
  booktitle={The Fourteenth International Conference on Learning Representations}
}

@article{nerfplayer,
  title={Nerfplayer: A streamable dynamic scene representation with decomposed neural radiance fields},
  author={Song, Liangchen and Chen, Anpei and Li, Zhong and Chen, Zhang and Chen, Lele and Yuan, Junsong and Xu, Yi and Geiger, Andreas},
  journal={IEEE Transactions on Visualization and Computer Graphics},
  volume={29},
  number={5},
  pages={2732--2742},
  year={2023},
  publisher={IEEE}
}

@InProceedings{hexplane,
    author    = {Cao, Ang and Johnson, Justin},
    title     = {{HexPlane}: A Fast Representation for Dynamic Scenes},
    booktitle = {Proceedings of the IEEE/CVF Conference on Computer Vision and Pattern Recognition (CVPR)},
    month     = {June},
    year      = {2023},
    pages     = {130-141}
}

@inproceedings{kplanes,
      title={K-planes: Explicit radiance fields in space, time, and appearance},
  author={Fridovich-Keil, Sara and Meanti, Giacomo and Warburg, Frederik Rahb{\ae}k and Recht, Benjamin and Kanazawa, Angjoo},
  booktitle={Proceedings of the IEEE/CVF conference on computer vision and pattern recognition},
  pages={12479--12488},
  year={2023}
}

@article{mixvoxels, title={Mixed neural voxels for fast multi-view video synthesis},
  author={Wang, Feng and Tan, Sinan and Li, Xinghang and Tian, Zeyue and Song, Yafei and Liu, Huaping},
  booktitle={Proceedings of the IEEE/CVF International Conference on Computer Vision},
  pages={19706--19716},
  year={2023}}

@inproceedings{swings, title={Swings: sliding windows for dynamic 3d gaussian splatting},
  author={Shaw, Richard and Nazarczuk, Michal and Song, Jifei and Moreau, Arthur and Catley-Chandar, Sibi and Dhamo, Helisa and P{\'e}rez-Pellitero, Eduardo},
  booktitle={European Conference on Computer Vision},
  pages={37--54},
  year={2024},
  organization={Springer}}

@InProceedings{neu3d, author = {Li, Tianye and Slavcheva, Mira and Zollh\"ofer, Michael and Green, Simon and Lassner, Christoph and Kim, Changil and Schmidt, Tanner and Lovegrove, Steven and Goesele, Michael and Newcombe, Richard and Lv, Zhaoyang}, title = {Neural 3D Video Synthesis From Multi-View Video}, booktitle={Proceedings of the IEEE/CVF conference on computer vision and pattern recognition}, month = {June}, year = {2022}, pages = {5521-5531} }

@inproceedings{timeformer,
  title={Timeformer: Capturing temporal relationships of deformable 3d gaussians for robust reconstruction},
  author={Jiang, Dadong and Hou, Zhi and Ke, Zhihui and Yang, Xianghui and Zhou, Xiaobo and Qiu, Tie},
  booktitle={Proceedings of the IEEE/CVF International Conference on Computer Vision},
  pages={8721--8732},
  year={2025}
}

@article{transformer,
  title={Attention is all you need},
  author={Vaswani, Ashish and Shazeer, Noam and Parmar, Niki and Uszkoreit, Jakob and Jones, Llion and Gomez, Aidan N and Kaiser, {\L}ukasz and Polosukhin, Illia},
  journal={Advances in neural information processing systems},
  volume={30},
  year={2017}
}

@InProceedings{spacetimegs,
  author    = {Li, Zhan and Chen, Zhang and Li, Zhong and Xu, Yi},
  title     = {Spacetime Gaussian Feature Splatting for Real-Time Dynamic View Synthesis},
  booktitle = {Proceedings of the IEEE/CVF Conference on Computer Vision and Pattern Recognition (CVPR)},
  month     = {June},
  year      = {2024},
  pages     = {8508--8520}
}

@inproceedings{st4dgs,
  author    = {Li, Deqi and Huang, Shi-Sheng and Lu, Zhiyuan and Duan, Xinran and Huang, Hua},
  title     = {{ST-4DGS}: Spatial-Temporally Consistent 4D Gaussian Splatting for Efficient Dynamic Scene Rendering},
  booktitle = {ACM SIGGRAPH 2024 Conference Papers},
  year      = {2024},
  doi       = {10.1145/3641519.3657520}
}

@InProceedings{freetimegs,
  author    = {Wang, Yifan and Yang, Peishan and Xu, Zhen and Sun, Jiaming and Zhang, Zhanhua and Chen, Yong and Bao, Hujun and Peng, Sida and Zhou, Xiaowei},
  title     = {{FreeTimeGS}: Free Gaussian Primitives at Anytime Anywhere for Dynamic Scene Reconstruction},
  booktitle = {Proceedings of the IEEE/CVF Conference on Computer Vision and Pattern Recognition (CVPR)},
  year      = {2025},
  pages     = {21750--21760}
}

@inproceedings{attal2023hyperreel,
  title={Hyperreel: High-fidelity 6-dof video with ray-conditioned sampling},
  author={Attal, Benjamin and Huang, Jia-Bin and Richardt, Christian and Zollhoefer, Michael and Kopf, Johannes and O’Toole, Matthew and Kim, Changil},
  booktitle={Proceedings of the IEEE/CVF Conference on Computer Vision and Pattern Recognition},
  pages={16610--16620},
  year={2023}
}


\clearpage
\appendix
\section*{Supplementary Material}
\addcontentsline{toc}{section}{Supplementary Material}
\setcounter{section}{0}
\renewcommand{\thesection}{\Alph{section}}

In this supplementary material, we provide additional implementation details, experiments and analyses to complement the main paper. This document is organized as follows:
\begin{itemize}
    \item \textbf{Sec.~\ref{sec:impl}}: Detailed implementation of training schedule, motion descriptor computation, and feature injection.
    \item \textbf{Sec.~\ref{sec:supp_ablation}}: Additional ablation and sensitivity analysis, covering motion score fusion weights, motion statistics update interval, temporal window size, temporal consistency, ablation variant details, and rendering efficiency.
    \item \textbf{Sec.~\ref{sec:plugin_validation}}: Plug-in validation beyond the DeGauss framework.
    \item \textbf{Sec.~\ref{sec:efficiency}}: Rendering efficiency and model compactness analysis.
    \item \textbf{Sec.~\ref{sec:limitations}}: Additional visual analysis, limitations, and future works.
\end{itemize}

\section{Detailed Implementation}\label{sec:impl}

We train all models using the Adam optimizer on a single NVIDIA RTX~3090 GPU. The three-stage training schedule is dataset-dependent: for Neu3D scenes, the coarse, fine, and post stages last 2k, 3k, and 20k iterations, respectively; for NeRF On-the-go scenes, the stages are 3k, 7k, and 20k iterations. At each motion-statistics update step, we construct a temporally ordered timestamp list from the training cameras and sample at most $K{=}64$ timestamps. The sampled deformation trajectories are further subsampled with stride~16 for computational efficiency. The resulting per-Gaussian motion descriptor is 13-dimensional and is cached to amortize computation. Before caching, the descriptor is normalized using exponential moving average (EMA) statistics with decay~0.99, a numerical stabilizer $\epsilon{=}10^{-6}$, and soft-clipping factor~3.0.

For feature injection, the motion descriptor is first encoded via sinusoidal positional encoding and then projected to the deformation hidden dimension through a two-layer MLP of width~256. The last linear layer is zero-initialized to ensure residual injection is identity at the beginning of training. Both the motion-statistics computation and the temporal feature construction are implemented using chunk-based processing over Gaussians, which keeps peak memory usage bounded and enables scaling to large dynamic scenes. During inference, cached motion statistics are directly reused without recomputation.

\section{Additional Ablation and Sensitivity Analysis}
\label{sec:supp_ablation}

This section provides additional ablation and sensitivity analyses for the key design components of MVFusion-GS, including motion score fusion weights, motion statistics update interval, temporal window size, temporal consistency, and detailed ablation variants.

\subsection{Sensitivity to Motion Score Fusion Weights}

In Eq.~(9), the motion intensity score is computed by fusing variance signals from Gaussian position, rotation, and scale attributes with default weights $0.7/0.2/0.1$, respectively. To validate this choice, we evaluate several alternative fusion strategies including uniform weighting, position-only variance, and partial attribute combinations. Results are summarized in Table~\ref{tab:weight_ablation}.

\begin{table}[h]
\centering
\caption{Sensitivity analysis of motion score fusion weights on the Neu3D dataset.}
\label{tab:weight_ablation}
\begin{tabular}{c c c c c c}
\toprule
Position & Rotation & Scale & PSNR $\uparrow$ & SSIM $\uparrow$ & LPIPS $\downarrow$ \\
\midrule
\textbf{0.7} & \textbf{0.2} & \textbf{0.1} & \textbf{32.07} & \textbf{0.943} & \textbf{0.046} \\
1/3 & 1/3 & 1/3 & 31.88 & 0.940 & 0.049 \\
1 & 0 & 0 & 31.72 & 0.936 & 0.051 \\
0.8 & 0.2 & 0 & 31.96 & 0.942 & 0.047 \\
0.8 & 0 & 0.2 & 31.91 & 0.940 & 0.047 \\
\bottomrule
\end{tabular}
\end{table}

\begin{figure}[h]
\centering
\includegraphics[width=\linewidth]{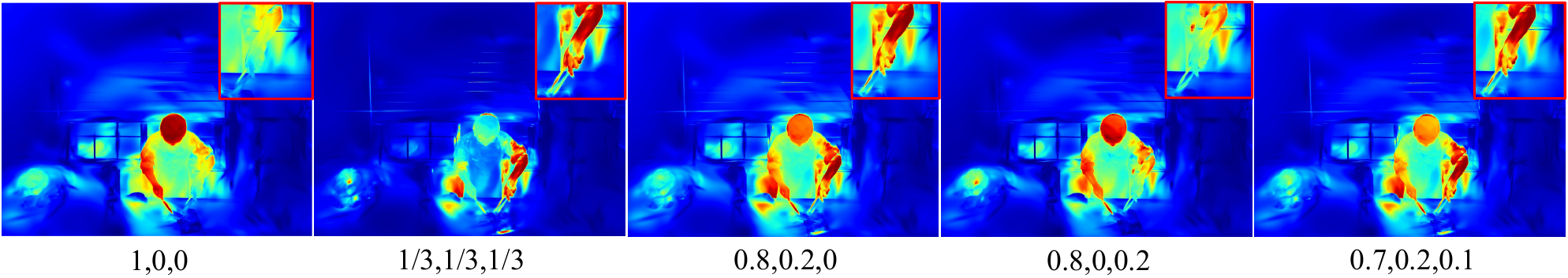}
\caption{
Visualization of motion variance maps under different fusion weight settings.
From left to right: position-only $(1,0,0)$, uniform weighting $(1/3,1/3,1/3)$, position+rotation $(0.8,0.2,0)$, position+scale $(0.8,0,0.2)$, and the proposed $(0.7,0.2,0.1)$.
Highlighted regions indicate areas with noticeable motion differences.
}
\label{fig:score_weight}
\end{figure}

Positional variance provides the dominant motion cue, yet relying on it alone leads to incomplete motion coverage. As illustrated in Fig.~\ref{fig:score_weight}, position-only variance highlights only the most salient moving regions while missing subtler motion areas. Incorporating rotation or scale variance individually improves spatial coverage but cannot fully capture all motion patterns; uniform weighting over-emphasizes the less reliable signals. The proposed asymmetric weighting $(0.7,0.2,0.1)$ achieves the best balance, preserving the strong motion sensitivity of positional variance while leveraging rotation and scale as complementary cues.

\subsection{Sensitivity to Motion Statistics Update Interval}

The motion statistics are periodically refreshed during training with an interval $E$, controlling how frequently the cached motion descriptors are updated from the current deformation predictions. We analyze the effect of different intervals on both Neu3D and NeRF On-the-go. Results are shown in Table~\ref{tab:update_interval}.

\begin{table}[h]
\centering
\caption{Sensitivity to motion statistics update interval $E$.}
\label{tab:update_interval}
\begin{tabular}{c ccc ccc}
\toprule
& \multicolumn{3}{c}{Neu3D} & \multicolumn{3}{c}{NeRF On-the-go} \\
\cmidrule(lr){2-4} \cmidrule(lr){5-7}
Update Interval $E$
& PSNR $\uparrow$ & SSIM $\uparrow$ & LPIPS $\downarrow$
& PSNR $\uparrow$ & SSIM $\uparrow$ & LPIPS $\downarrow$ \\
\midrule
100  & 31.89 & 0.937 & 0.053 & 23.06 & 0.812 & 0.092 \\
500  & 32.02 & 0.940 & 0.049 & 23.31 & 0.819 & 0.089 \\
1000 & 32.05 & 0.941 & 0.047 & 23.91 & 0.823 & 0.086 \\
\textbf{2000} & \textbf{32.07} & \textbf{0.943} & \textbf{0.046} & \textbf{23.94} & \textbf{0.826} & \textbf{0.085} \\
\bottomrule
\end{tabular}
\end{table}

Overly frequent updates introduce noisy motion estimates during early training when the deformation field is still evolving, while excessively sparse updates produce stale descriptors that lag behind newly learned motion patterns. As shown in Table~\ref{tab:update_interval}, $E{=}2000$ achieves the best trade-off between stability and adaptability on both benchmarks.

\subsection{Sensitivity to Temporal Window Size}

The temporal window size $w$ determines how many neighboring timestamps are used by MFTA to estimate local motion context. We evaluate different window sizes on both Neu3D and NeRF On-the-go. Results are reported in Table~\ref{tab:window}.

\begin{table}[h]
\centering
\caption{Sensitivity to temporal window size $w$.}
\label{tab:window}
\begin{tabular}{c ccc ccc}
\toprule
& \multicolumn{3}{c}{Neu3D} & \multicolumn{3}{c}{NeRF On-the-go} \\
\cmidrule(lr){2-4} \cmidrule(lr){5-7}
Window Size $w$
& PSNR $\uparrow$ & SSIM $\uparrow$ & LPIPS $\downarrow$
& PSNR $\uparrow$ & SSIM $\uparrow$ & LPIPS $\downarrow$ \\
\midrule
3 & 31.98 & 0.938 & 0.052 & 23.35 & 0.816 & 0.096 \\
\textbf{5} & \textbf{32.07} & \textbf{0.943} & \textbf{0.046} & \textbf{23.94} & \textbf{0.826} & \textbf{0.085} \\
7 & 32.06 & 0.941 & 0.047 & 23.69 & 0.820 & 0.089 \\
9 & 32.04 & 0.940 & 0.049 & 23.60 & 0.819 & 0.092 \\
\bottomrule
\end{tabular}
\end{table}

A small window provides insufficient temporal context, leading to unstable motion estimation. Increasing the window improves robustness by aggregating information from neighboring timestamps; however, overly large windows introduce irrelevant temporal variations that dilute the motion signal. A moderate window size of $w{=}5$ yields the most reliable motion estimation on both benchmarks.

\subsection{Temporal Consistency and Simpler Alternatives}
\label{sec:temporal_alternatives}

We further evaluate whether the improvement comes from motion-variance-guided temporal modeling rather than generic temporal smoothing. We report T-LPIPS on Neu3D to measure temporal stability and compare with several controlled alternatives.

\begin{table}[h]
\centering
\caption{Temporal consistency and controlled alternatives on Neu3D. Lower T-LPIPS indicates better temporal stability.}
\label{tab:temporal_ablation}
{
\begin{tabular}{lcc}
\toprule
Method & PSNR $\uparrow$ & T-LPIPS $\downarrow$ \\
\midrule
4DGS~\cite{4dgs} & 31.12 & 0.115 \\
DeGauss~\cite{degauss} & 31.52 & 0.099 \\
MangoGS~\cite{mangogs} & 31.85 & 0.092 \\
w/o MFTA & 31.78 & 0.090 \\
Direct deformation magnitude & unstable & unstable \\
Temporal averaging & 31.74 & 0.092 \\
Temporal convolution & 31.81 & 0.090 \\
Ours & \textbf{32.07} & \textbf{0.081} \\
\bottomrule
\end{tabular}}
\end{table}

Removing MFTA worsens both PSNR and T-LPIPS, confirming the contribution of temporal attention. Temporal averaging and temporal convolution improve the baseline but remain below MVFusion-GS, suggesting that the gain is not simply due to smoothing neighboring frames. Directly injecting deformation magnitude failed to converge stably because the raw deformation values have large ranges and high dimensionality, making optimization harder.

\subsection{Ablation Variant Details}
\label{sec:ablation_detail}

For completeness, we summarize the implementation of each ablation variant used in Table 5 in the main paper.

\begin{enumerate}
\item \textbf{w/o MVG \& MFTA (Base deformation).}
Both modules are removed. The model reduces to the baseline deformation network that predicts per-Gaussian motion directly from spatial coordinates and timestamps.

\item \textbf{w/o MVG (no motion statistics).}
The motion-statistics branch in MVG is removed, so no trajectory-level variance descriptor is injected into the deformation feature. The remaining refinement pipeline is kept unchanged.

\item \textbf{w/ MVG (position variance only).}
Only positional variance is used as the motion statistic, while rotation and scale variances are excluded from the descriptor.

\item \textbf{w/o MFTA.}
The MFTA module is removed while MVG remains active, disabling temporal attention but retaining motion statistics.

\item \textbf{w/o Cross-attention (Self-attention).}
The query-centered cross-attention mechanism in MFTA is replaced with self-attention over temporal tokens.

\item \textbf{Full Model.}
The complete model combines the full MVG descriptor with MFTA temporal attention.
\end{enumerate}

\section{Plug-in Validation Beyond DeGauss}
\label{sec:plugin_validation}

To examine whether the proposed modules are specific to the DeGauss decomposition framework, we integrate MVG and MFTA into the standard 4DGS pipeline without adding static-dynamic decomposition. This setting keeps the original unified deformation-based 4DGS pipeline and only augments its deformation feature with the proposed motion-aware refinement.

\begin{table}[h]
\centering
\small
\caption{Plug-in validation on Neu3D. Adding MVG and MFTA to the standard 4DGS pipeline improves reconstruction quality without introducing static-dynamic decomposition.}
\label{tab:plugin_validation}
\begin{tabular}{lccc}
\toprule
Method & PSNR $\uparrow$ & SSIM $\uparrow$ & LPIPS $\downarrow$ \\
\midrule
4DGS~\cite{4dgs} & 31.12 & 0.937 & 0.058 \\
4DGS+MVG+MFTA & \textbf{31.66} & \textbf{0.942} & \textbf{0.057} \\
\bottomrule
\end{tabular}
\end{table}

The plug-in result shows a 0.54 dB PSNR improvement over 4DGS, indicating that the proposed motion-aware refinement is not limited to the DeGauss decomposition framework. Since the modules only rely on deformation features and predicted motion statistics, they can also benefit standard deformation-based dynamic Gaussian pipelines.

\section{Rendering Efficiency and Model Compactness}
\label{sec:efficiency}

We evaluate the rendering efficiency of our method and compare it with representative dynamic Gaussian approaches. Following standard practice, rendering speed is measured in frames per second (FPS) at a fixed resolution on a single GPU. Results are reported in Table~\ref{tab:quality_efficiency}.

\begin{table*}[h]
  \centering
  \caption{Quality and efficiency evaluation on all scenes of Neu3D~\cite{neu3d} tested on an RTX~4090. Dyn.\ GS Num denotes the number of dynamic Gaussian primitives in the final model.}
  \label{tab:quality_efficiency}
  {%
  \begin{tabular}{l|ccc|ccc}
    \toprule
    Method & PSNR$\uparrow$ & SSIM$\uparrow$ & LPIPS$\downarrow$ & Training Time$\downarrow$ & FPS$\uparrow$ & Dyn.\ GS Num$\downarrow$ \\
    \midrule
    NeRFPlayer~\cite{nerfplayer}
      & 30.29 & 0.909 & 0.151 & 6 hours & 0.045 & - \\
    HyperReel~\cite{attal2023hyperreel}
      & 30.72 & 0.931 & 0.101 & 9 hours & 2.0 & - \\
    HexPlane~\cite{hexplane}
      & 30.00 & 0.922 & 0.113 & 12 hours & 0.2 & - \\
    KPlanes~\cite{kplanes}
      & 31.63 & \textbf{0.963} & 0.117 & 5.0 hours & 0.3 & - \\
    SWinGS~\cite{swings}
      & 31.12 & 0.941 & 0.095 & - & 71 & - \\
    \midrule
    4DGS~\cite{4dgs}
      & 31.12 & 0.937 & 0.058 & 0.85 hours & 53.1 & 124,197 \\
    DeGauss~\cite{degauss}
      & 31.52 & 0.942 & 0.047 & 2.1 hours & 71.0 & 56,533 \\
    \midrule
    Ours  & \textbf{32.07} & 0.943 & \textbf{0.046} & 2.26 hours & 62.7 & 32,985 \\
    \bottomrule
  \end{tabular}%
  }
\end{table*}

Our method achieves the highest PSNR and lowest LPIPS on the Neu3D benchmark while maintaining a competitive rendering speed of 62.7~FPS. Although slightly slower than DeGauss due to the additional motion-aware computation, our model significantly reduces the number of dynamic Gaussians (32,985 vs.\ 56,533), indicating a more compact scene representation.

\section{Additional Visual Analysis and Limitations}
\label{sec:limitations}

\subsection{Branch-Level Decomposition Visualization}

\begin{figure}[h]
    \centering
    \includegraphics[width=0.95\linewidth]{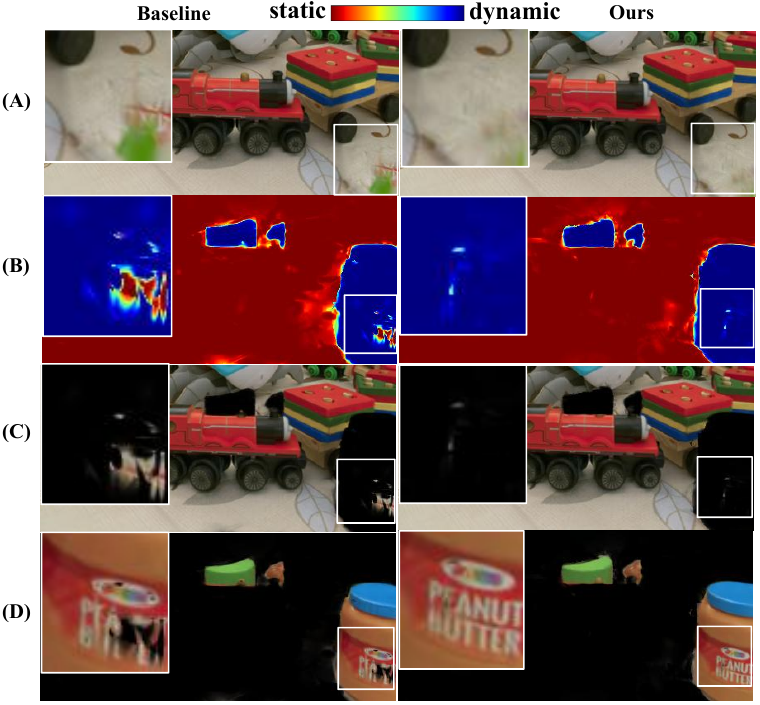}
    \caption{\textbf{Branch-level decomposition visualization.}
    The baseline without MVG/MFTA retains pseudo-static foreground residuals in the static branch, causing visible foreground artifacts. MVFusion-GS suppresses these residuals and better explains the foreground through the dynamic branch, showing that cleaner background reconstruction comes from strengthened motion-aware deformation rather than post-hoc background correction.
    (A) Failure Case Static Background; (B) Pred Mask; (C) Masked Static Background; (D) Masked Dynamic Foreground.}
    \label{fig:branch_vis}
\end{figure}

Fig.~\ref{fig:branch_vis} provides a branch-level visualization of the dynamic-static decomposition. Compared with the baseline, MVFusion-GS reduces foreground-like residuals in the static branch and assigns the corresponding content to the dynamic branch. This supports our main observation that improving the deformation branch helps prevent ambiguous moving content from being absorbed into the static reconstruction.

\subsection{Failure Cases and Future Work}

\begin{figure}[t]
    \centering
    \includegraphics[width=\linewidth]{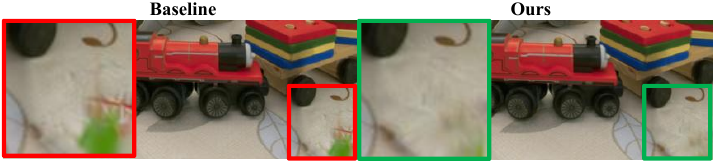}
    \caption{\textbf{Failure case.}
    In challenging cases where foreground motion is complex, the estimated deformation variance may become less discriminative, especially when the baseline deformation field underfits foreground motion. MVFusion-GS can substantially reduce pseudo-static foreground residuals, but it may not completely remove artifacts when the underlying deformation model fails to provide reliable motion cues.}
    \label{fig:failure}
\end{figure}

Although motion variance statistics effectively guide dynamic-static separation, they still depend on the baseline deformation field to provide meaningful motion cues. When the initial deformation is severely under-fitted, affected by heavy occlusion, or contains extremely weak motion evidence, the cached variance prior may become less discriminative and residual artifacts can remain, as shown in Fig.~\ref{fig:failure}. A promising direction for future work is to leverage these variance cues for explicit Gaussian pruning, reassignment, or dynamic-static partition refinement. Another interesting direction is to couple motion-variance estimation with stronger deformation priors or uncertainty-aware temporal modeling, so that ambiguous low-motion regions can be handled more robustly during early training.
\end{document}